\documentclass[manuscript,screen, noacm]{acmart}

\usepackage{amsmath}
\usepackage{graphicx}
\usepackage{hyperref}
\usepackage[utf8]{inputenc}
\usepackage{natbib}
\usepackage{xcolor}
\usepackage{ifthen}
\usepackage[normalem]{ulem}
\usepackage{svg}
\usepackage{subfiles}
\usepackage{xspace}
\usepackage{cleveref}
\usepackage{wrapfig}
\usepackage{dialogue}
\usepackage[most]{tcolorbox}
\hypersetup{ hidelinks }

\acmConference[FAccT '22]{ACM Conference on Fairness, Accountability, and Transparency 2022}{June 21--24, 2022}{Seoul, South Korea}

\AtBeginDocument{%
  \providecommand\BibTeX{{%
    \normalfont B\kern-0.5em{\scshape i\kern-0.25em b}\kern-0.8em\TeX}}}

\usepackage{pifont}

\copyrightyear{2022}
\acmYear{2022}
\setcopyright{rightsretained}
\acmConference[FAccT '22]{2022 ACM Conference on Fairness, Accountability, and Transparency}{June 21--24, 2022}{South Korea}
\acmBooktitle{2022 ACM Conference on Fairness, Accountability, and Transparency (FAccT '22)}
\acmDOI{}
\acmISBN{}
\setcopyright{none}
\renewcommand\footnotetextcopyrightpermission[1]{} 
\begin{document}

\title{The Values Encoded in Machine Learning Research}

\author{Abeba Birhane}
  \authornote{All authors contributed equally to this research.}
\email{abeba@mozillafoundation.org}
\orcid{0000-0001-6319-7937}
\affiliation{%
  \institution{Mozilla Foundation \& School of Computer Science, University College Dublin}
  \streetaddress{Belfied}
  \city{Dublin}
  \country{Ireland}
}
\author{Pratyusha Kalluri*}
\email{pkalluri@stanford.edu}
\orcid{0000-0001-7202-8027}
\affiliation{
  \institution{Computer Science Department, Stanford University}
  \streetaddress{353 Jane Stanford Way}
  \city{Palo Alto}
  \country{USA}
}
\author{Dallas Card*}
\email{dalc@umich.edu}
\orcid{0000-0001-5573-8836}
\affiliation{ 
  \institution{School of Information, University of Michigan}
  \streetaddress{105 S State St}
  \city{Ann Arbor}
  \country{USA}
}
\author{William Agnew*}
\email{wagnew3@cs.washington.edu}
\orcid{}
\affiliation{%
  \institution{Paul G. Allen School of Computer Science and Engineering, University of Washington}
  \streetaddress{185 E Stevens Way NE}
  \city{Seattle}
  \country{USA}
}
\author{Ravit Dotan*}
\email{ravit.dotan@berkeley.edu}
\orcid{0000-0002-9646-8315}
\affiliation{
  \institution{Center for Philosophy of Science, University of Pittsburgh}
  \streetaddress{4200 Fifth Ave}
  \city{Pittsburgh}
  \country{USA}
}
\author{Michelle Bao*}
\email{baom@stanford.edu}
\orcid{0000-0002-4410-0703}
\affiliation{%
  \institution{Computer Science Department, Stanford University}
  \streetaddress{353 Jane Stanford Way}
  \city{Palo Alto}
  \country{USA}
}

\newcommand{\rk}[1]{\textcolor{blue}{#1}}
\newcommand{\teal}[1]{\textcolor{teal}{#1}}
\newcommand{\pink}[1]{\textcolor{magenta}{#1}}
\def\b{{\raisebox{.13ex}{\footnotesize{\ding{70}\hspace{4pt}}}}}

\newcommand{\beginemph}[1]{\noindent
\textbf{\textit{{#1}}}~}
\newcommand{\specialemph}[1]{\textbf{{{#1}}}}

\begin{abstract}

Machine learning currently exerts an outsized influence on the world, increasingly affecting institutional practices and impacted communities. It is therefore critical that we question vague conceptions of the field as value-neutral or universally beneficial, and investigate what specific values the field is advancing. In this paper, we first introduce a method and annotation scheme for studying the values encoded in documents such as research papers. Applying the scheme, we analyze 100 highly cited machine learning papers published at premier machine learning conferences, ICML and NeurIPS. We annotate key features of papers which reveal their values: their justification for their choice of project, which attributes of their project they uplift, their consideration of potential negative consequences, and their institutional affiliations and funding sources. We find that few of the papers justify how their project connects to a societal need (15\%) and far fewer discuss negative potential (1\%). Through line-by-line content analysis, we identify 59 values that are uplifted in ML research, and, of these, we find that the papers most frequently justify and assess themselves based on Performance, Generalization, Quantitative evidence, Efficiency, Building on past work, and Novelty. 
We present extensive textual evidence and identify key themes in the definitions and operationalization of these values. Notably, we find systematic textual evidence that these top values are being defined and applied with assumptions and implications generally supporting the centralization of power.
Finally, we find increasingly close ties between these highly cited papers and tech companies and elite universities.

\end{abstract}

\maketitle

\section{Introduction}

Over recent decades, machine learning (ML) has risen from a relatively obscure research area to an extremely influential discipline, actively being deployed in myriad applications and contexts around the world. 
Current discussions of ML frequently follow a historical strain of thinking which has tended to frame technology as "neutral", based on the notion that new technologies can be unpredictably applied for both beneficial and harmful purposes \citep{winner.1977}. This claim of neutrality frequently serves as an insulation from critiques of AI and as permission to emphasize the benefits of AI \citep{rus.2018, weizenbaum1972impact, nanayakkara.2021}, often without any acknowledgment that benefits and harms are distributed unevenly.
Although it is rare to see anyone explicitly argue in print that ML is neutral, related ideas are part of contemporary conversation, including these canonical claims: long term impacts are too difficult to predict; sociological impacts are outside the expertise or purview of ML researchers \citep{holstein.2019}; critiques of AI are really misdirected critiques of those deploying AI with bad data ("garbage in, garbage out"),  again outside the purview of many AI researchers; and proposals such as broader impact statements represent merely a "bureaucratic constraint"
\citep{abuhamad.2020}. ML research is often cast as value-neutral and emphasis is placed on positive applications or potentials. Yet,
the objectives and values of ML research are influenced by many social forces that shape factors including what research gets done and who benefits.\footnote{For example, ML research is influenced by social factors including the personal preferences of researchers and reviewers, other work in science and engineering, the interests of academic institutions, funding agencies and companies, and larger systemic pressures, including systems of oppression.} Therefore, it is important to challenge perceptions of neutrality and universal benefit, and document and understand the emergent values of the field: what specifically the field is prioritizing and working toward. To this end, we perform an in-depth analysis of 100 highly cited NeurIPS and ICML papers from four recent years.

Our key contributions are as follows:

\begin{enumerate}
\item \textbf{We present and open source a fine-grained annotation scheme for the study of values in documents such as research papers.}\footnote{\label{footnote:annotation}
We include our annotation scheme and all annotations at
\href{https://github.com/wagnew3/The-Values-Encoded-in-Machine-Learning-Research}{https://github.com/wagnew3/The-Values-Encoded-in-Machine-Learning-Research} with a CC BY-NC-SA license.}
To our knowledge, our annotation scheme is the first of its kind and opens the door to further qualitative and quantitative analyses of research.
This is a timely methodological contribution, as institutions including prestigious ML venues and community organizations are increasingly seeking and reflexively conducting interdisciplinary study on social aspects of machine learning \citep{blodgett2020language, bender2021dangers, lewis2018digital, bengio2021retrospective}.

\item \textbf{We apply our scheme to annotate 100 influential ML research papers and extract their value commitments, including identifying 59 values significant in machine learning research.} These papers reflect and shape the values of the field. Like the annotation scheme, the resulting repository of over 3,500 annotated sentences is available and is valuable as foundation for further qualitative and quantitative study.

\item \textbf{We perform extensive textual analysis to understand dominant values}: Performance, Generalization, Efficiency, Building on past work, and Novelty.
Our analysis reveals that while these values may seem on their face to be purely technical, they are 
socially and politically charged: \textbf{we find systematic textual evidence corroborating that these values are currently defined and operationalized in ways that centralize power}, i.e., disproportionally benefit and empower the already powerful, while neglecting society's least advantaged.\footnote{We understand this to be an interdisciplinary contribution: Scholarship on the values of ML (or alternatives) often faces dismissal based on perceived distance from prestigious ML research and quantifiable results. Meanwhile, philosophers of science have been working to understand the roles and political underpinnings of values in science for decades, e.g., in biology and social sciences \citep{kuhn.1977, longino.1996}. Our paper provides convincing qualitative and quantitative evidence of ML values and their political underpinnings, bridging ML research and both bodies of work.}

\item \textbf{We present a quantitative analysis of the affiliations and funding sources of these influential papers. We find substantive and increasing presence of tech corporations.} For example, in 2008/09, 24\% of these top cited papers had corporate affiliated authors, and in 2018/19 this statistic more than doubled, to 55\%. Moreover, of these corporations connected to influential papers, the presence of "big-tech" firms, such as Google and Microsoft, more than tripled from 21\% to 66\%.
\end{enumerate}

\section{Methodology}

To study the values of ML research, we conduct an in-depth analysis of ML research papers distinctively informative of these values.\footnote{Because the aim of qualitative inquiry is depth of understanding, it is viewed as important to 
analyze information-rich documents (those that distinctively reflect and shape the central values of machine learning; for example, textual analysis of influential papers) in lieu of random sampling and broad analysis (for example, keyword frequencies in a large random sample of ML papers). This is referred to as the importance of purposive sampling \citep{patton1990qualitative}.} We chose to focus on highly cited papers because they reflect and shape the values of the discipline, drawing from NeurIPS and ICML because they are the most prestigious of the long-running ML conferences.\footnote{At the time of writing, NeurIPS and ICML, along with the newer conference ICLR,
comprised the top 3 conferences
according to h5-index (and h5-median) in the AI category on Google Scholar, by a large margin. Citation counts are based on the Semantic Scholar database.} Acceptance to these conferences is a valuable commodity used to evaluate researchers, and submitted papers are typically explicitly written so as to win the approval of the community, particularly the reviewers who will be drawn from that community. As such, these papers effectively reveal the values that authors believe are most valued by that community. Citations indicate amplification by the community, and help to position these papers as influential exemplars of ML research. To avoid detecting only short-lived trends, we drew papers from two recent years (2018/19\footnote{At the time of beginning annotation, 2018 and 2019 were the two most recent years available.}) and from ten years earlier (2008/09).
We focused on conference papers because they tend to follow a standard format and 
allow limited space, meaning that researchers must make hard choices about what to emphasize.
Collectively, an interdisciplinary team of researchers analyzed the 100 most highly cited papers from NeurIPS and ICML, from the years 2008, 2009, 2018, and 2019, annotating over 3,500 sentences drawn from them. In the context of expert  content analysis, this constitutes a large scale annotation which allows us to meaningfully comment on central values.

\begin{figure*}
\centering
\includegraphics[origin=c,width=0.75\linewidth]{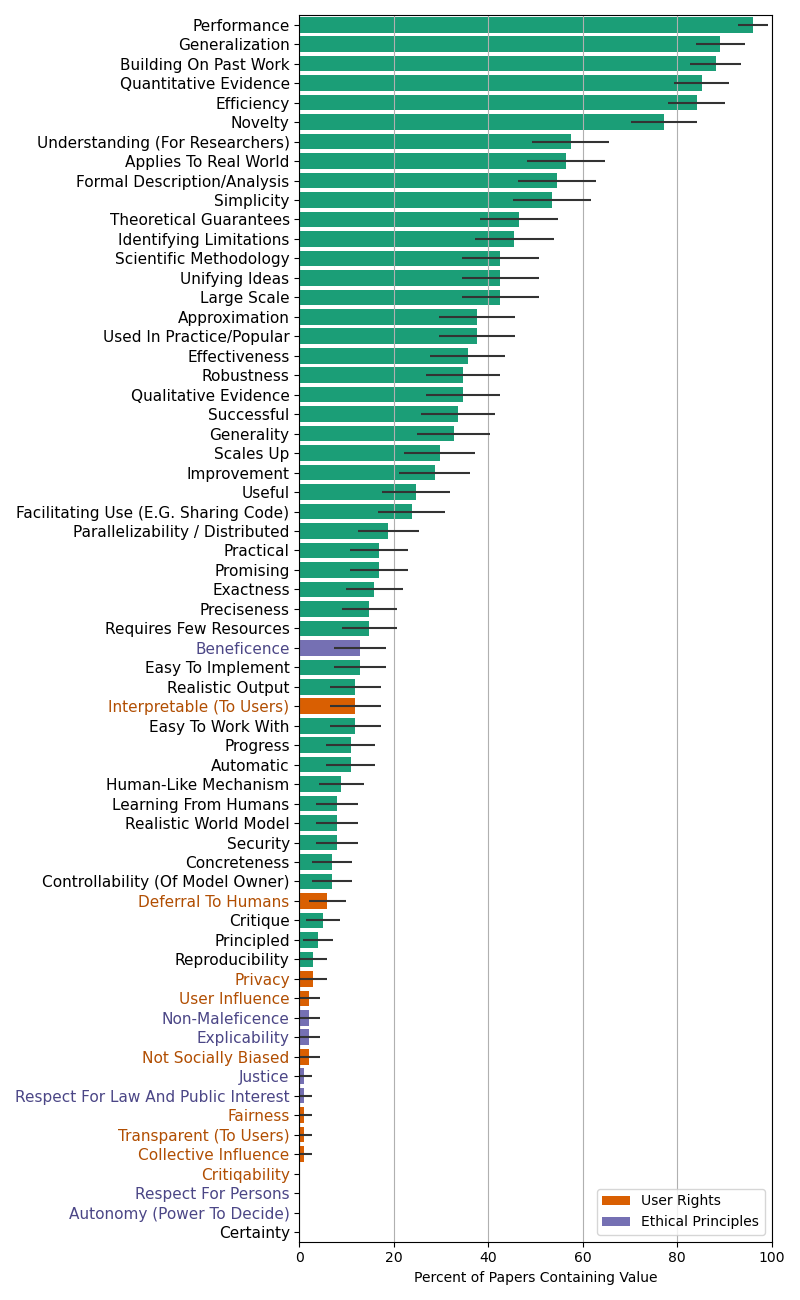}
\caption{Proportion of annotated papers that uplift each value.}
\label{fig:value-totals}
\end{figure*}

Our team constructed an annotation scheme and applied it to manually annotate each paper, examining the abstract, introduction, discussion, and conclusion: (1) We examined the chain of reasoning by which each paper justified its contributions, which we call the \textit{justificatory chain}, categorizing the extent to which papers used technical or societal problems to justify or motivate their contributions (Table~\ref{tab:just_chain}).\footnote{In qualitative research, the term `coding' is used to denote deductively categorizing text into selected categories as well as inductively annotating text with emergent categories. To avoid overloading computer science `coding', we use the terms categorizing and annotating throughout this paper.}\textsuperscript{,}\footnote{\label{footnote:rigor}We found the first three categories of this scheme 
were generally sufficient for our analysis. 
In service of rich understanding, we included the subtler fourth category.
As much as possible, we steel-manned discussions: regardless of whether we were convinced or intrigued by a discussion, if it presented the level of detail typical when discussing projects' technical implications, then it was assigned category four.} (2) We carefully read each sentence of these sections line-by-line, inductively annotating any and all values uplifted by the sentence (Figure~\ref{fig:value-totals}). We use a conceptualization of "value" that is widespread in philosophy of science in theorizing about values in sciences: a "value" of an entity is a property that is considered desirable for that kind of entity, e.g. regarded as a desirable attribute for machine learning research.\footnote{For example, speed can be described as valuable in an antelope \cite{mcmullin1982values}. Well-know scientific values include accuracy, consistency, scope, simplicity, and fruitfulness \cite{kuhn.1977}. See \cite{longino.1996} for a critical discussion of socially-laden aspects of these values in science.} (3) We categorized the extent to which the paper included a discussion of potential negative impacts (Table~\ref{tab:neg_impacts}).\textsuperscript{\ref{footnote:rigor}} (4) We documented and categorized the author affiliations and stated funding sources. 
In this paper, we provide complete annotations,
quantize the annotations to quantify and present dominant patterns, and present randomly sampled excerpts and key themes in how these values become socially loaded.

To perform the line-by-line analysis and annotate the uplifted values (Figure~\ref{fig:value-totals}), we used a hybrid inductive-deductive content analysis methodology and followed best practices \citep{hsieh2005three, merriam2019qualitative, bengtsson.2016,krippendorff.2018}:
(i) We began with several values of interest based on prior literature, specifically seven ethical principles and user rights
\citep{dittrich.2012,floridi.2019, kalluri2019values}.
(ii) We randomly sampled a subset of 10 papers for initial annotation, reading sentence by sentence, deductively annotating for the values of interest and inductively adding new values as they emerged, by discussion until perfect consensus. 
The deductive component ensures we note and can speak to values of interest, and the inductive component enables discovery and impedes findings limited by bias or preconception by requiring textual grounding and focusing on emergent values
\citep{bengtsson.2016,krippendorff.2018}.
(iii) We annotated the full set of papers sentence by sentence. We followed the constant comparative method, in which we continually compared each text unit to the annotations and values list thus far, annotated for the values in the values list, held regular discussions, and we individually nominated and decided by consensus when sentences required inductively adding emergent values to the values list \citep{glaser1999discovery}. 
We used a number of established strategies in service of consistency which we discuss below.
Following qualitative research best practices, we identified by consensus a small number of values we found were used synonymously or closely related and combined these categories, listing all merges in Appendix \ref{sec:value_clusters}.\footnote{For example, in Section~\ref{sec:efficiency}, we discuss themes cutting across efficiency, sometimes referenced in the abstract and sometimes indicated by uplifting data efficiency, energy efficiency, fast, label efficiency, low cost, memory efficiency, or reduced training time.}
(iv) In this paper, for each top value, we present randomly selected quotations of the value, richly describe the meaning of the value in context, 
present key themes in how the value is operationalized and becomes socially loaded, and illustrate its contingency by comparing to alternative values in the literature that might have been or might be valued instead.

We adhere to a number of best practices to establish reliability: We practice prolonged engagement, conducting long-term orientation to and analysis of data over more than a year (in lieu of short-term analysis that is dominated by preconceptions) \citep{lincoln2006naturalistic}; We triangulate across researchers (six researchers) and points in time (four years) and place (two conferences) \citep{patton1999enhancing, denzin2017sociological};
We recode data coded early in the process \citep{krefting.1991}; We transparently publish the complete annotation scheme and all annotations \citep{noble2015issues}; We conduct negative case analysis, for example, drawing out and discussing papers with unusually strong connections to societal needs \citep{lincoln2006naturalistic}; and we include a reflexivity statement in Appendix \ref{app:reflexivity}
describing our team in greater detail, striving to highlight relevant personal and disciplinary viewpoints. 

The composition of our team confers additional validity to our work. We are a multi-racial, multi-gender team working closely, including undergraduate, graduate, and post-graduate researchers engaged with machine learning, NLP, robotics, cognitive science, critical theory, community organizing, and philosophy. This team captures several advantages: the nature of this team minimizes personal and intra-disciplinary biases, affords the unique combination of expertise required to read the values in complex ML papers, allows meaningful engagement with relevant work in other fields, and enabled best practices including continually clarifying the procedure, ensuring agreement, vetting consistency, reannotating, and discussing themes \citep{krippendorff.2018}. Across the annotating team, we found that annotators were able to make somewhat different and complementary inductive paper-level observations, while obtaining near or perfect consensus on corpus-level findings. To assess the consistency of paper-level annotations, 40\% of the papers were double-annotated by paired annotators. During the inductive-deductive process of annotating sentences with values (ultimately annotating each sentence for the presence of 75 values), paired annotators agreed 87.0\% of the time, and obtained a fuzzy Fleiss’ kappa \cite{kirilenko2016inter} on values per paper of 0.45, indicating moderate agreement. During the deductive process of categorizing the extent to which a paper included societal justification and negative potential impacts (ordinal categorization according to the schema in Table~\ref{tab:just_chain} and Table~\ref{tab:neg_impacts}), paired annotators obtained substantial agreement, indicated by Fleiss’ weighted kappa ($\kappa$=.60, $\kappa$=.79). Finally, at the corpus level we found substantial agreement: annotators identified the list of emergent values by perfect consensus, unanimously finding these values to be present in the papers. Across annotators, there was substantial agreement on the relative prevalence (ranking) of the values, indicated by Kendall’s \textit{W} \cite{kendall1939problem} (W=.80), and we identified by consensus the five most dominant values, which we discuss in detail.

Manual analysis is necessary at all steps of the method (i-iv). Manual analysis is required for the central task of reading the papers and inductively identifying previously unobserved values. Additionally, once values have been established, we find manual analysis continues to be necessary for annotation. We find that many values are expressed in ways that are subtle, varied, or rely on contextual knowledge. We find current automated methods for labeling including keyword searches and basic classifiers miss new values, annotate poorly relative to manual annotation, and systematically skew the results towards values which are easy to identify, while missing or mischaracterizing values which are exhibited in more nuanced ways.\footnote{In Appendix \ref{app:automatic}, we implement automatic annotation and empirically demonstrate these failure modes.
}  Accordingly, we find our use of qualitative methodology is indispensable. Reading all papers is key for contributing the textual analysis as well, as doing so includes developing a subtle understanding of how the values function in the text and understanding of taken for granted assumptions underlying the values.

In the context of an interdisciplinary readership, including ML and other STEM disciplines that foreground quantitative methodology, it is both a unique contribution and a limitation that this paper centers qualitative methodology. Ours is a significant and timely methodological contribution as there is rising interest in qualitatively studying the social values being encoded in ML, including reflexively by ML researchers \citep{blodgett2020language, bender2021dangers, lewis2018digital, bengio2021retrospective}. Simultaneously, the use of qualitative methodology in quantitative-leaning contexts could lead to misinterpretations. Human beliefs are complex and multitudinous, and it is well-established that when qualitative-leaning methodology is presented in quantitative-leaning contexts, it is possible for study of imprecise subject matter to be misinterpreted as imprecise study of subjects \citep{berg2017qualitative}.

In brief, whereas quantitative analysis typically favors large random sampling and strict, statistical evidence in service of generalization of findings, qualitative analysis typically favors purposive sampling from information-rich context and richly descriptive evidence in service of depth of understanding \citep{merriam2019qualitative, berg2017qualitative}. 
For both our final list of values and specific annotation of individual sentences, different researchers might make somewhat different choices.  However, given the overwhelming presence of certain values, the high agreement rate among annotators, and the similarity of observations made by our team, we believe other researchers following a similar approach would reach similar conclusions about what values are most frequently uplifted. Also, we cannot claim to have identified every relevant value in ML. Rather, we present a collection of such values; and by including important ethical values identified by past work, and specifically looking for these, we can confidently assert their relative absence in this set of papers.
Finally, qualitative analysis is an effort to understand situations in their uniqueness, i.e., in this set of papers. Future work may determine whether and how to form conclusions about  stratifications (e.g. between chosen years or conferences) and whether and how to use this qualitative analysis to construct new quantitative instruments to ascertain generalization (e.g. across more years or conferences) \citep{patton1990qualitative, doyle2020overview}. Our study contributes unprecedent data and textual analysis and lays the groundwork for this future work.

\section{Quantitative Summary}
\label{sec:quant}

In Figure~\ref{fig:value-totals}, we plot the prevalence of values in 100 annotated papers. The top values are: performance (96\% of papers), generalization (89\%),
building on past work (88\%), quantitative evidence (85\%), efficiency (84\%), and novelty (77\%). Values related to user rights and stated in ethical principles appeared very rarely if at all: none of the papers mentioned autonomy, justice, or respect for persons.
In Table~\ref{tab:just_chain}, we show the distribution of justification scores. Most papers only justify how they achieve their internal, technical goal; 68\% make no mention of societal need or impact, and only 4\% make a rigorous attempt to present links connecting their research to societal needs.
In Table~\ref{tab:neg_impacts}, we show the distribution of negative impact discussion scores. One annotated paper included a discussion of negative impacts and a second mentioned the possibility of negative impacts. 98\% of papers contained no reference to potential negative impacts.
In Figure~\ref{fig:vcorp-ties}, we show stated connections (funding ties and author affiliations) to  institutions. Comparing papers written in 2008/2009 to those written in 2018/2019, ties to corporations nearly doubled to 79\% of all annotated papers, ties to big tech more than tripled, to 66\%, while ties to universities declined to 81\%, putting the presence of corporations nearly on par with universities. In the next section, we present extensive qualitative examples and analysis of our findings.

\begin{table*}
\caption{Annotations of justificatory chain.}\label{tab:just_chain}
\centering
\small
\begin{tabular}{lc}
\toprule
\textbf{Justificatory Chain} & \textbf{\% of Papers} \\ \midrule
Does not mention societal need & 68\% \\
States but does not justify how it connects to a societal need & 17\% \\
States and somewhat justifies how it connects to a societal need & 11\% \\
States and rigorously justifies how it connects to a a societal need & 4\% \\ \bottomrule
\end{tabular}%
\end{table*}%

\begin{table*}
\caption{Annotations of discussed negative potential.}\label{tab:neg_impacts}
\centering
\small
\begin{tabular}{lc}
\toprule
\textbf{Discussion of Negative Potential} & \textbf{\% of Papers} \\ \midrule
Does not mention negative potential & 98\% \\
Mentions but does not discuss negative potential \hspace{18 mm} & 1\% \\
Discusses negative potential & 1\% \\
Deepens our understanding of negative potential & 0\% \\ \bottomrule
\end{tabular}%
\end{table*}%

\section{Textual analysis}
\label{sec:qual}

\subsection{Justifications}

We find papers typically justify their choice of project by contextualizing it within a broader goal and giving a chain of justification from the broader goal to the particular project pursued in the paper. These justifications reveal priorities:

\specialemph{Papers typically motivate their projects by appealing to the needs of the ML research community and rarely mention potential societal benefits.} Research-driven needs of the ML community include researcher understanding (e.g., understanding the effect of pre-training on performance/robustness, theoretically understanding multi-layer networks) as well as more practical research problems (e.g., improving efficiency of models for large datasets, creating a new benchmark for NLP tasks). 

\specialemph{Even when societal needs are mentioned as part of the justification of the project, the connection is loose.} Some papers do appeal to needs of  broader society, such as building models with realistic assumptions, catering to more languages, or ``understanding the world''. Yet almost no papers explain how their project promotes a social need they identify by giving the kind of rigorous justification that is typically expected of and given for technical contributions.

\specialemph{The cursory nature of the connection between societal needs and the content of the paper also manifests in the fact that the societal needs, or the applicability to the real world, is often only discussed in the beginning of the papers.} From papers that mention applicability to the real world, the vast majority of mentions are in the Introduction section, and applicability is rarely engaged with afterwards. Papers tend to introduce the problem as useful for applications in object detection or text classification, for example, but rarely justify why an application is worth contributing to, or revisit how they particularly contribute to an application as their result.

\subsection{Discussion of Negative Potential}
\label{negative potential}
Although a plethora of work exists on sources of harm that can arise in relation to ML research \cite{buolamwini2018gender, green2019good, suresh2019framework, hill2020accused,bender2021dangers}, we observe that these discussions are ignored in these influential conference publications.

\specialemph{It is extremely rare for papers to mention negative potential at all.} 
Just as the goals of the papers are largely inward-looking, prioritizing the needs of the ML research community, these papers fail to acknowledge both broader societal needs and societal impacts. This norm is taken for granted: none of these papers offer any explanation for why they cannot speak to negative impacts. These observations correspond to a larger trend in the ML research community of neglecting to discuss aspects of the work that are not strictly positive. 

\specialemph{The lack of discussion of potential harms is especially striking for papers which deal with contentious application areas}, such as surveillance and misinformation. These include papers, for example, that advance identification of people in images, face-swapping, and video synthesis. These papers contain no mention of the well-studied negative potential of facial surveillance, DeepFakes, or misleading videos. 

\specialemph{Among the two papers that do mention negative potential, 
the discussions were mostly
abstract and hypothetical},
rather than grounded in the concrete negative potential of their specific contributions. For example, authors may acknowledge "possible unwanted social biases" when applying models to a real-world setting, without 
commenting on let alone assessing the social biases encoded in the authors' proposed model.

\subsection{Stated values}
\label{sec:values}

The dominant values that emerged from the annotated corpus are: Performance, Generalization, Building on past work, Quantitative evidence, Efficiency, and Novelty. These are often portrayed as innate and purely technical. However, the following analysis of these values shows how they can become politically loaded in the process of prioritizing and operationalizing them: sensitivity to the way that they are operationalized, and to the fact that they are uplifted at all, reveals value-laden assumptions that are often taken for granted.
To provide a sense of what the values look like in context, Tables \ref{tab:performance}, \ref{tab:generalization}, \ref{tab:efficiency} and \ref{tab:novelty_building} present randomly selected examples of sentences annotated with the values of Performance, Generalization, Efficiency, Building on past work, and Novelty respectively.
Extensive additional examples can be found in Appendix \ref{sec:randomexamples}.\footnote{To avoid the impression that we are mainly interested in drawing attention to specific papers, we omit attribution for individual examples, but include a list of all annotated papers in Appendix \ref{sec:paperlist}. Note that most sentences are annotated with multiple values; for example, there can be overlap in sentences annotated with \textit{performance} and sentences annotated with \textit{generalization}.
}
For each of these prominent values, we quantify its dominance, identify constituent values that contribute to this value, challenge a conception of the value as politically neutral, identify key themes in how the value is socialy loaded, and we cite alternatives to its dominant conceptualization that may be equally or more valid, interesting, or socially beneficial.
When values seem neutral or innate, we have encouraged ourselves, and now encourage the reader, to remember that values once held to be intrinsic, obvious, or definitional have in many cases been found harmful and transformed over time and purportedly neutral values warrant careful consideration.

\subsection{Performance}

\begin{table}
\caption{Random examples of \textit{performance}, the most common emergent value.}
\label{tab:performance}
\small
\begin{tabular}{p{.96\linewidth}}
\midrule "Our model significantly outperforms SVM’s, and it also outperforms convolutional neural nets when given additional unlabeled data produced by small translations of the training images."\\
\midrule "We show in simulations on synthetic examples and on the IEDB MHC-I binding dataset, that our approach outperforms well-known convex methods for multi-task learning, as well as related non-convex methods dedicated to the same problem."\\
\midrule "Furthermore, the learning accuracy and performance of our LGP approach will be compared with other important standard methods in Section 4, e.g., LWPR [8], standard GPR [1], sparse online Gaussian process regression (OGP) [5] and $\upsilon$-support vector regression ($\upsilon$-SVR) [11], respectively."\\ 
\midrule "In addition to having theoretically sound grounds, the proposed method also outperformed state-of-the-art methods in two experiments with real data."\\
\midrule "We prove that unlabeled data bridges this gap: a simple semisupervised learning procedure (self-training) achieves high robust accuracy using the same number of labels required for achieving high standard accuracy."
\\
 \midrule
"Experiments show that PointCNN achieves on par or better performance than state-of-the-art methods on multiple challenging benchmark datasets and tasks."
 \\ 
 \midrule 
"Despite its impressive empirical performance, NAS is computationally expensive and time consuming, e.g. Zoph et al. (2018) use 450 GPUs for 3-4 days (i.e. 32,400-43,200 GPU hours)."
\\ 
 \midrule 
"However, it is worth examining why this combination of priors results in superior performance."
 \\ 
 \midrule 
 "In comparisons with a number of prior HRL methods, we find that our approach substantially outperforms previous state-of-the-art techniques."
 \\ 
 \midrule 
 "Our proposed method addresses these issues, and greatly outperforms the current state of the art."
 \\
 \bottomrule
\end{tabular}\label{fig}
\end{table}
Emphasizing performance is the most common way by which papers attempt to communicate their contributions, by showing a specific, quantitative, improvement over past work, according to some metric on a new or established dataset. For some reviewers, obtaining better performance than any other system---a ``state-of-the-art'' (SOTA) result---is seen as a noteworthy, or even necessary, contribution \citep{rogers.2020}. 

Despite acknowledged issues with this kind of evaluation (including the artificiality of many datasets, and the privileging of ``tricks'' over insight; \citealp{lipton.2018,ethayarajh.2020}), performance is typically presented as intrinsic to the field. Frequently, the value of Performance is indicated by specifically uplifting accuracy or state of the art results, which are presented as similarly intrinsic. However, models are not simply "well-performing" or "accurate" in the abstract but always in relation to and as quantified by some \textit{metric} on some \textit{dataset}. Examining definition and  operationalization of performance values, we identify three key social aspects. 

\b \specialemph{Performance values are consistently and without discussion operationalized as correctness averaged across individual predictions, giving equal weight to each instance.} However, choosing
equal weights when averaging is a value-laden move which might deprioritize those underrepresentated in the data or the world, as well as societal and evaluee needs and preferences regarding inclusion. Extensive research in ML fairness and related fields has considered alternatives, but we found no such discussions among the influential papers we examined.

\b \specialemph{Datasets are typically preestablished, large corpora with discrete "ground truth" labels. }They are often driven purely by past work, so as to demonstrate improvement over a previous baseline (see also \S\ref{sec:novelty}). Another common justification for using a certain dataset is claimed applicability to the "real world". Assumptions about how to characterize the "real world" are value-laden. One preestablished and typically perpetuated assumption is the availability of very large datasets. However, presupposing the availability of large datasets is non-neutral and power centralizing because it encodes favoritism to those with resources to obtain and process them \cite{dotan2019value}. Additionally, the welfare, consent, or awareness of the datafied subjects whose images end up in a large scale image dataset, for example, are not considered in the annotated papers. Further overlooked assumptions include that the real world is binary or discrete, and that datasets come with a predefined ground-truth label for each example, presuming that a true label always exists "out there" independent of those carving it out, defining and labelling it. This contrasts against marginalized scholars’ calls for ML models that allow for non-binaries, plural truths, contextual truths, and many ways of being \cite{costanza2018design, hamidi2018gender, lewis2020indigenous}.

\b \specialemph{The prioritization of performance values is so entrenched in the field that generic success terms, such as "success", "progress", or "improvement" are used as synonyms for performance and accuracy.} However, one might alternatively invoke generic success to mean increasingly safe, consensual, or participatory ML that reckons with impacted communities and the environment. In fact, "performance" itself is a general success term that could have been associated with properties other than accuracy and SOTA.

\subsection{Generalization}

\begin{table}
\caption{Random examples of \textit{generalization}, the second most common emergent value.}
\label{tab:generalization}
\small
\begin{tabular}
{p{0.96\linewidth}}
\midrule "The range of applications that come with generative models are vast, where audio synthesis [55] and semi-supervised classification [38, 31, 44] are examples hereof."\\
\midrule "Furthermore, the infinite limit could conceivably make sense in deep learning, since over-parametrization seems to help optimization a lot and doesn’t hurt generalization much [Zhang et al., 2017]: deep neural nets with millions of parameters work well even for datasets with 50k training examples."\\
\midrule "Combining the optimization and generalization results, we uncover a broad class of learnable functions, including linear functions, two-layer neural networks with polynomial activation $\phi(z) = z^{2l}$ or cosine activation, etc."\\
\midrule "We can apply the proposed method to solve regularized least square problems, which have the loss function $(1 - y_i\omega^T x_i)^2$ in (1)."\\
\midrule "The result is a generalized deflation procedure that typically outperforms more standard techniques on real-world datasets."\\
\midrule "Our proposed invariance measure is broadly applicable to evaluating many deep learning algorithms for many tasks, but the present paper will focus on two different algorithms applied to computer vision."\\
\midrule "We show how both multitask learning and semi-supervised learning improve the generalization of the shared tasks, resulting in state-of-the-art performance." \\
\midrule "We have also demonstrated that the proposed model is able to generalize much better than LDA in terms of both the log-probability on held-out documents and the retrieval accuracy." \\
\midrule "We define a rather general convolutional network architecture and describe its application to many well known NLP tasks including part-of-speech tagging, chunking, named-entity recognition, learning a language modeland the task of semantic role-labeling" \\
\midrule "We demonstrate our algorithm on multiple datasets and show that it outperforms relevant baselines."\\

\bottomrule
\end{tabular}
\hfill

\label{fig}
\end{table}

We observe that a common way of appraising the merits of one's work is to claim that it generalizes well. Notably, generalization is understood in terms of the dominant value, performance: a model is perceived as generalizing when it achieves good performance on a range of samples, datasets, domains, tasks, or applications. 
In fact, the value of generalization is sometimes indicated by referencing generalization in the abstract and other times indicated by specifically uplifting values such as Minimal discrepancy between train/test samples or Flexibility/extensibility, e.g., to other tasks. 
We identify three key socially loaded aspects of how generalization is defined and operationalized.

\b \specialemph{Only certain datasets, domains, or applications are valued as indicators of model generalization.} Typically, a paper shows that a model generalizes by showing that it performs well on multiple tasks or datasets. However, like the tasks and datasets indicating performance, the choice of particular tasks and datasets indicating generalization is rarely justified; the choice of tasks can often seem arbitrary, and authors often claim generalization while rarely presenting discussion or analysis indicating their results will generalize outside the carefully selected datasets, domains or applications, or to more realistic settings, or help to directly address societal needs.

\b \specialemph{Prizing generalization leads institutions to harvest datasets from various domains, and to treat these as the only datasets that matter in the space of problems}. Papers prizing generalization implicitly and sometimes explicitly prioritize reducing every scenario top-down to a common set of representations or affordances, rather than treating each setting as meaningfully unique and potentially motivating technologies or lack thereof that are fundamentally different from the current standard. Despite vague associations between generalization and accessible technology for diverse peoples, in practice work on generalization frequently targets one model to rule them all, denigrating diverse access needs. Critical scholars have advocated for valuing \emph{context}, which may stand opposed to striving for generalization \citep{d2020data}. Others have argued that this kind of totalizing lens (in which model developers have unlimited power to determine how the world is represented) leads to \emph{representational} harms, due to applying a single representational framework to everything \citep{crawford.2017,abbasi.2019}.

\b \specialemph{The belief that generalization is possible assumes new data will be or should be treated similarly to previously seen data. }
When used in the context of ML, the assumption that the future resembles the past is often problematic as past societal stereotypes and injustice can be encoded in the process \cite{o2016weapons}. Furthermore, to the extent that predictions are performative \cite{perdomo2020performative}, especially predictions that are enacted, those ML models which are deployed to the world will contribute to shaping social patterns. None of the annotated papers attempt to counteract this quality or acknowledge its presence.

\subsection{Efficiency}
\label{sec:efficiency}

\begin{table}
\caption{Random examples of \textit{efficiency}, the fifth most common emergent value.}
\label{tab:efficiency}
\begin{small}
\begin{tabular}{p{0.96\linewidth}}
\midrule "Our model allows for controllable yet efficient generation of an entire news article – not just the body, but also the title, news source, publication date, and author list."\\
\midrule "We show that Bayesian PMF models can be efficiently trained using Markov chain Monte Carlo methods by applying them to the Netflix dataset, which consists of over 100 million movie ratings."\\
\midrule "In particular, our EfficientNet-B7 surpasses the best existing GPipe accuracy (Huang et al., 2018), but using 8.4x fewer parameters and running 6.1x faster on inference."\\
\midrule "Our method improves over both online and batch methods and learns faster on a dozen NLP datasets."\\
\midrule “We describe efficient algorithms for projecting a vector onto the $\ell$1-ball.”\\
\midrule "Approximation of this prior structure through simple, efficient hyperparameter optimization steps is sufficient to achieve these performance gains."\\
\midrule "We have developed a new distributed agent IMPALA (Importance Weighted Actor-Learner Architecture) that not only uses resources more efficiently in single-machine training but also scales to thousands of machines without sacrificing data efficiency or resource utilisation."\\
\midrule "In this paper we propose a simple and efficient algorithm SVP (Singular Value Projection) based on the projected gradient algorithm"\\
\midrule "We give an exact and efficient dynamic programming algorithm to compute CNTKs for ReLU activation." \\
\midrule "In contrast, our proposed algorithm has strong bounds, requires no extra work for enforcing positive definiteness, and can be implemented efficiently." \\

\bottomrule
\end{tabular}
\hfill

\end{small}
\label{fig}
\end{table}

In the annotated papers, we find that saying that a model is efficient typically indicates the model uses less of some resource, e.g., data efficiency, energy efficiency, label efficiency, memory efficiency, being low cost, fast, or having reduced training time. 
We find that the definition and operationalization of efficiency encodes key social priorities, namely  \textit{which kind of efficiency matters} and \textit{to what end}.  

\b \specialemph{Efficiency is commonly referenced to indicate the ability to scale up, not to save resources.} For example, a more efficient inference method allows you to do inference in much larger models or on larger datasets, using the same amount of resources used previously, or more.
This mirrors the classic Jevon's paradox: greater resource efficiency often leads to overall greater utilization of that resource.
This is reflected in our value annotations, where 84\% of papers mention valuing efficiency, but only 15\% of those value requiring \textit{few resources}. When referencing the consequences of efficiency, many papers present evidence that efficiency enables scaling up, while none of the papers present evidence that efficiency can facilitate work by low-resource communities or can lessen resource extraction -- e.g. less hardware or data harvesting or lower carbon emissions.
In this way, valuing efficiency facilitates and encourages the most powerful actors to scale up their computation to ever higher orders of magnitude, making their models even less accessible to those without resources to use them and decreasing the ability to compete with them. Alternative usages of efficiency could encode accessibility instead of scalability, aiming to create more equitable conditions.

\subsection{Novelty and Building on Past Work}
\label{sec:novelty}

\begin{table}
\caption{Random examples of \emph{building on past work} and \emph{novelty}, the third and sixth most common emergent values, respectively.}
\label{tab:novelty_building}
\begin{small}
\begin{tabular}{p{0.96\linewidth}}
\toprule
\textbf{Building on past work}\\
\midrule "Recent work points towards sample complexity as a possible reason for the small gains in robustness: Schmidt et al. [41] show that in a simple model, learning a classifier with non-trivial adversarially robust accuracy requires substantially more samples than achieving good `standard' accuracy."\\
\midrule "Experiments indicate that our method is much faster than state of the art solvers such as Pegasos, TRON, SVMperf, and a recent primal coordinate descent implementation."\\
\midrule "There is a large literature on GP (response surface) optimization."\\
\midrule "In a recent breakthrough, Recht et al. [24] gave the first nontrivial results for the problem obtaining guaranteed rank minimization for affine transformations A that satisfy a restricted isometry property (RIP)."\\
\midrule "In this paper, we combine the basic idea behind both approaches, i.e., LWPR and GPR, attempting to get as close as possible to the speed of local learning while having a comparable accuracy to Gaussian process regression"\\
\bottomrule
\toprule
\textbf{Novelty} \\
\midrule "In this paper, we propose a video-to-video synthesis approach under the generative adversarial learning framework."\\
\midrule "Third, we propose a novel method for the listwise approach, which we call ListMLE."\\
\midrule "The distinguishing feature of our work is the use of Markov chain Monte Carlo (MCMC) methods for approximate inference in this model."\\
\midrule "To our knowledge, this is the first attack algorithm proposed for this threat model."\\
\midrule "Here, we focus on a different type of structure, namely output sparsity, which is not addressed in previous work."\\
\bottomrule
\end{tabular}
\hfill

\end{small}
\label{fig:novelty}
\end{table}

Most authors devote space in the introduction to positioning their paper in relation to past work, and describing what is novel.
Building on past work is sometimes referenced broadly and other times is indicated more specifically as building on classic work or building on recent work.
In general, mentioning past work serves to signal awareness of related publications, 
to establish the new work as relevant to the community, and to provide the basis upon which to make claims about what is new. Novelty is sometimes suggested implicitly (e.g., "we develop" or "we propose"), but frequently it is emphasized explicitly (e.g. "a new algorithm" or "a novel approach").
The emphasis on novelty is common across many academic fields \citep{trapido.2015,vinkers.2015}.
The combined focus on novelty and building on past work establishes a continuity of ideas, and might be expected to contribute to the self-correcting nature of science \citep{merton.1973}. However, this is not always the case \citep{ioannidis.2012} and attention to the ways novelty and building on past work are defined and implemented reveals two key social commitments. 

\b \specialemph{Technical novelty is most highly valued.} The highly-cited papers we examined mostly tend to emphasize the novelty of their proposed method or of their theoretical result. Very few uplifted their paper on the basis of applying an existing method to a novel domain, or for providing a novel philosophical argument or synthesis. We find a clear emphasis on technical novelty, rather than critique of past work, or demonstration of measurable progress on societal problems, as has previously been observed \citep{wagstaff.2012}. 

\b \specialemph{Although introductions sometimes point out limitations of past work so as to further emphasize the contributions of their own paper, they are rarely explicitly critical of other papers in terms of datasets, methods, or goals.} Indeed, papers uncritically reuse the same datasets for years or decades to benchmark their algorithms, even if those datasets fail to represent more realistic contexts in which their algorithms will be used \cite{bender2021dangers}. Novelty is denied to work that critiques or rectifies socially harmful aspects of existing datasets and goals, and this occurs in tandem with strong pressure to benchmark on them and thereby perpetuate their use, enforcing a conservative bent to ML research.

\section{Corporate Affiliations and Funding}
\label{sec:corporate}

\begin{figure}
\centering
\captionsetup{justification=centering}
\includegraphics[width=0.7\linewidth]{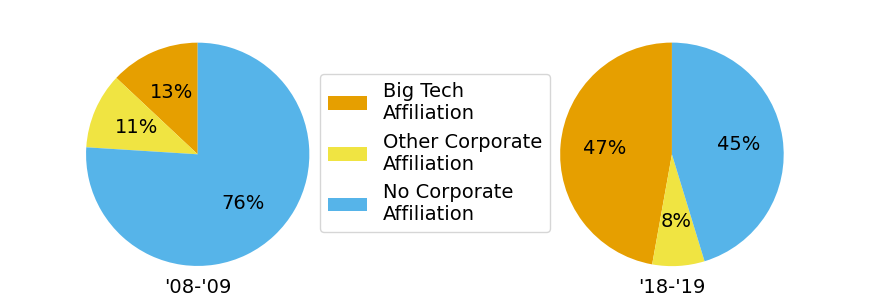}
\caption{Corporate and Big Tech author affiliations. \\ The percent of papers with Big Tech author affiliations increased from 13\% in 2008/09 to 47\% in 2018/19.
}
\label{fig:corp_affils}
\vspace*{-0.25cm}
\end{figure}

\begin{figure*}
\centering
\captionsetup{justification=centering}
\includegraphics[width=1.0\linewidth]{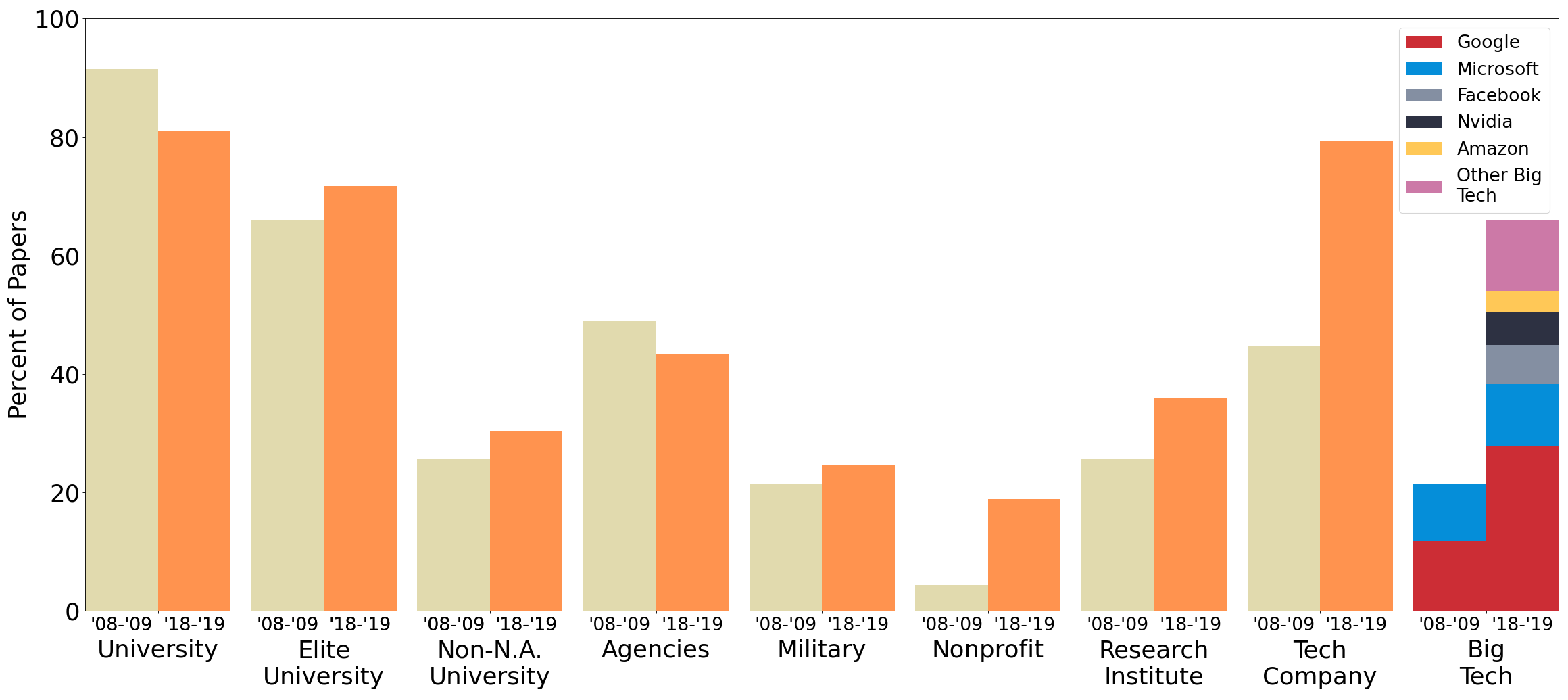}
\caption{Affiliations and funding ties. \\ From 2008/09 to 2018/19, the percent of papers tied to nonprofits, research institutes, and tech companies increased substantially. Most significantly, ties to Big Tech increased threefold and overall ties to tech companies increased to 79\%. \\ Non-N.A. Universities are those outside the U.S. and Canada.}
\label{fig:vcorp-ties}
\vspace*{-0.25cm}
\end{figure*}

\beginemph{Quantitative summary.~} Our analysis shows substantive and increasing corporate presence in the most highly-cited papers.
In 2008/09, 24\% of the top cited papers had \textit{corporate affiliated authors}, and in 2018/19 this statistic more than doubled to 55\%. Furthermore, we also find a much greater concentration of a few large tech firms, such as Google and Microsoft, with the presence of these "big tech" firms (as identified in \citep{ahmed2020democratization})
increasing nearly fourfold, from 13\% to 47\% (Figure \ref{fig:corp_affils}). The fraction of the annotated papers with corporate ties by \textit{corporate affiliated authors or corporate funding} dramatically increased from 45\% in 2008/09 to 79\% in 2018/19 (Figure \ref{fig:vcorp-ties}).
These findings are consistent with contemporary work indicating a pronounced corporate presence in ML research: in an automated analysis of
peer-reviewed papers from 57 major computer science conferences, Ahmed and Wahed \cite{ahmed2020democratization} show that the share of papers with corporate affiliated authors increased from 10\% in 2005 for both ICML and NeurIPS to 30\% and 35\% respectively in 2019. Our analysis shows that corporate presence is even more pronounced in those papers from ICML and NeurIPS that end up receiving the most citations. In addition, we found paramount domination of elite universities in our analysis as shown in Figure \ref{fig:vcorp-ties}. Of the total papers with university affiliations, we found 80\% were from elite universities (defined as the top 50 universities by QS World  University Rankings, following past work \cite{ahmed2020democratization}).

\beginemph{Analysis.~}
The influence of powerful players in ML research is consistent with field-wide value commitments that centralize power. Others have  argued for causal connections. For example, Abdalla and Abdalla \cite{abdalla2020grey} argue that big tech sway and influence academic and public discourse using strategies which closely resemble strategies used by Big Tabacco.
Moreover, examining the prevalent values of big tech, critiques have repeatedly pointed out that objectives such as efficiency, scale, and wealth accumulation \cite{o2016weapons,pasquale2015black,hanna2020against} drive the industry at large, often at the expense of individuals rights, respect for persons, consideration of negative impacts, beneficence, and justice. Thus, the top stated values of ML that we presented in this paper such as performance, generalization, and efficiency may not only enable and facilitate the realization of big tech's objectives, but also suppress values such as beneficence, justice, and inclusion. A "state-of-the-art" large image dataset, for example, is instrumental for large scale models, further benefiting ML researchers and big tech in possession of huge computing power. 
In the current climate — where values such as accuracy, efficiency, and scale, as currently defined, are a priority, and there is a pattern of centralization of power — user safety, informed consent, or participation may be perceived as costly and time consuming, evading social needs.

\section{Discussion and Related Work}

There is a foundational understanding in Science, Technology, and Society Studies (STS), Critical Theory, and Philosophy of Science that science and technologies are inherently value-laden, and these values are encoded in technological artifacts, many times in contrast to a field's formal research criteria, espoused consequences, or ethics guidelines \cite{winner1980artifacts,bowker2000sorting,benjamin2019race}. There is a long tradition of exposing and critiquing such values in technology and computer science. For example, Winner \cite {winner1980artifacts} introduced several ways technology can encode political values. This work is closely related to Rogaway \cite {rogaway2015moral}, who notes that cryptography has political and moral dimensions and argues for a cryptography that better addresses societal needs. 

Our paper extends these critiques to the field of ML. It is a part of a rich space of interdisciplinary critiques and alternative lenses used to examine the field. Works such as \cite {mohamed2020decolonial, birhane2020algorithmic} critique AI, ML, and data using a decolonial lens, noting how these technologies replicate colonial power relationships and values, and propose decolonial values and methods.  
Others \cite {benjamin2019race, noble2018algorithms, d2020data} examine technology and data science from an anti-racist and intersectional feminist lens, discussing how our infrastructure has largely been built by and for white men; D’Ignazio and Klein \cite {d2020data} present a set of alternative principles and methodologies for an intersectional feminist data science. Similarly, Kalluri \cite {kalluri2020don} denotes that the core values of ML are closely aligned with the values of the most privileged and outlines a vision where ML models are used to shift power from the most to the least powerful. Dotan and Milli \cite{dotan2019value} argue that the rise of deep learning is value-laden, promoting the centralization of power among other political values. Many researchers, as well as organizations such as Data for Black Lives, the Algorithmic Justice League, Our Data Bodies, the Radical AI Network, Indigenous AI, Black in AI, and Queer in AI, explicitly work on continuing to uncover particular ways technology in general and ML in particular can encode and amplify racist, sexist, queerphobic, transphobic, and otherwise marginalizing values, while simultaneously working to actualize alternatives \cite{buolamwini2018gender, prabhu2020large}.

There has been considerable growth over the past few years in institutional, academic, and grassroots interest in the societal impacts of ML, as reflected in the rise of relevant grassroots and non-profit organizations, the organizing of new workshops, the emergence of new conferences such as FAccT, and changes to community norms, such as the required broader impacts statements at NeurIPS. 
We present this paper in part to make visible the present state of the field and to demonstrate its contingent nature; it could be otherwise. 
For individuals, communities, and institutions wading through difficult-to-pin-down values of the field, as well as those striving toward alternative values, it is advantageous to have a characterization of the way the field is now — to serve as both a confirmation and a map for understanding, shaping, dismantling, or transforming what is, and for
articulating and 
bringing about alternative visions.

\section{Conclusion}

In this study, we find robust evidence against the vague conceptualization of the discipline of ML as value-neutral. Instead, we investigate the ways that the discipline of ML is inherently value-laden. Our analysis of highly influential papers in the discipline finds that they not only favor the needs of research communities and large firms over broader social needs, but also that they take this favoritism for granted, not acknowledging critiques or alternatives. The favoritism manifests in the choice of projects, the lack of consideration of potential negative impacts, and the prioritization and operationalization of values such as performance, generalization, efficiency, and novelty. These values are operationalized in ways that disfavor societal needs.
Moreover, we uncover an  overwhelming and increasing presence of big tech and elite universities in these highly cited papers, which is consistent with a system of power-centralizing value-commitments. 
The upshot is that the discipline of ML is not value-neutral. We present extensive quantitative and qualitative evidence that it is socially and politically loaded, frequently neglecting societal needs and harms, while prioritizing and promoting 
the concentration of resources, tools, knowledge, and power in the hands of already powerful actors.

 \begin{acks}

We would like to thank Luke Stark, Dan Jurafsky, and Sarah K. Dreier for helpful feedback on this work. We owe gratitude and accountability to the long history of work exposing how technology shifts power, work primarily done by communities at the margins. Abeba Birhane was supported in part by Science Foundation Ireland grant 13/RC/2094\_2. Pratyusha Kalluri was supported in part by the Open Phil AI Fellowship. Dallas Card was supported in part by the Stanford Data Science Institute. William Agnew was supported by an NDSEG Fellowship.

\end{acks}

\Urlmuskip=0mu plus 1mu
\bibliographystyle{ACM-Reference-Format}
\bibliography{FAccT}


\begin{thebibliography}{66}


\ifx \showCODEN    \undefined \def \showCODEN     #1{\unskip}     \fi
\ifx \showDOI      \undefined \def \showDOI       #1{#1}\fi
\ifx \showISBNx    \undefined \def \showISBNx     #1{\unskip}     \fi
\ifx \showISBNxiii \undefined \def \showISBNxiii  #1{\unskip}     \fi
\ifx \showISSN     \undefined \def \showISSN      #1{\unskip}     \fi
\ifx \showLCCN     \undefined \def \showLCCN      #1{\unskip}     \fi
\ifx \shownote     \undefined \def \shownote      #1{#1}          \fi
\ifx \showarticletitle \undefined \def \showarticletitle #1{#1}   \fi
\ifx \showURL      \undefined \def \showURL       {\relax}        \fi
\providecommand\bibfield[2]{#2}
\providecommand\bibinfo[2]{#2}
\providecommand\natexlab[1]{#1}
\providecommand\showeprint[2][]{arXiv:#2}

\bibitem[Abbasi et~al\mbox{.}(2019)]%
        {abbasi.2019}
\bibfield{author}{\bibinfo{person}{Mohsen Abbasi}, \bibinfo{person}{Sorelle~A.
  Friedler}, \bibinfo{person}{Carlos Scheidegger}, {and}
  \bibinfo{person}{Suresh Venkatasubramanian}.}
  \bibinfo{year}{2019}\natexlab{}.
\newblock \showarticletitle{Fairness in Representation: Quantifying
  Stereotyping as a Representational Harm}. In
  \bibinfo{booktitle}{\emph{Proceedings of the 2019 SIAM International
  Conference on Data Mining}}.
\newblock


\bibitem[Abdalla and Abdalla(2021)]%
        {abdalla2020grey}
\bibfield{author}{\bibinfo{person}{Mohamed Abdalla} {and}
  \bibinfo{person}{Moustafa Abdalla}.} \bibinfo{year}{2021}\natexlab{}.
\newblock \showarticletitle{The Grey Hoodie Project: {B}ig Tobacco, Big Tech,
  and the Threat on Academic Integrity}. In
  \bibinfo{booktitle}{\emph{Proceedings of the 2021 AAAI/ACM Conference on AI,
  Ethics, and Society}}.
\newblock
\urldef\tempurl%
\url{https://doi.org/10.1145/3461702.3462563}
\showURL{%
\tempurl}


\bibitem[Abuhamad and Rheault(2020)]%
        {abuhamad.2020}
\bibfield{author}{\bibinfo{person}{Grace Abuhamad} {and}
  \bibinfo{person}{Claudel Rheault}.} \bibinfo{year}{2020}\natexlab{}.
\newblock \showarticletitle{Like a Researcher Stating Broader Impact For the
  Very First Time}.
\newblock \bibinfo{journal}{\emph{arXiv preprint arXiv:2011.13032}}
  (\bibinfo{year}{2020}).
\newblock


\bibitem[Ahmed and Wahed(2020)]%
        {ahmed2020democratization}
\bibfield{author}{\bibinfo{person}{Nur Ahmed} {and} \bibinfo{person}{Muntasir
  Wahed}.} \bibinfo{year}{2020}\natexlab{}.
\newblock \showarticletitle{The De-democratization of {AI}: {Deep} Learning and
  the Compute Divide in Artificial Intelligence Research}.
\newblock \bibinfo{journal}{\emph{arXiv preprint arXiv:2010.15581}}
  (\bibinfo{year}{2020}).
\newblock


\bibitem[Ammar et~al\mbox{.}(2018)]%
        {ammar.2018}
\bibfield{author}{\bibinfo{person}{Waleed Ammar}, \bibinfo{person}{Dirk
  Groeneveld}, \bibinfo{person}{Chandra Bhagavatula}, \bibinfo{person}{Iz
  Beltagy}, \bibinfo{person}{Miles Crawford}, \bibinfo{person}{Doug Downey},
  \bibinfo{person}{Jason Dunkelberger}, \bibinfo{person}{Ahmed Elgohary},
  \bibinfo{person}{Sergey Feldman}, \bibinfo{person}{Vu Ha},
  \bibinfo{person}{Rodney Kinney}, \bibinfo{person}{Sebastian Kohlmeier},
  \bibinfo{person}{Kyle Lo}, \bibinfo{person}{Tyler Murray},
  \bibinfo{person}{Hsu-Han Ooi}, \bibinfo{person}{Matthew Peters},
  \bibinfo{person}{Joanna Power}, \bibinfo{person}{Sam Skjonsberg},
  \bibinfo{person}{Lucy~Lu Wang}, \bibinfo{person}{Chris Wilhelm},
  \bibinfo{person}{Zheng Yuan}, \bibinfo{person}{Madeleine van Zuylen}, {and}
  \bibinfo{person}{Oren Etzioni}.} \bibinfo{year}{2018}\natexlab{}.
\newblock \showarticletitle{Construction of the Literature Graph in Semantic
  Scholar}. In \bibinfo{booktitle}{\emph{Proceedings of NAACL}}.
\newblock
\urldef\tempurl%
\url{https://www.semanticscholar.org/paper/09e3cf5704bcb16e6657f6ceed70e93373a54618}
\showURL{%
\tempurl}


\bibitem[Bailey et~al\mbox{.}(2012)]%
        {dittrich.2012}
\bibfield{author}{\bibinfo{person}{Michael Bailey}, \bibinfo{person}{David
  Dittrich}, \bibinfo{person}{Erin Kenneally}, {and} \bibinfo{person}{Doug
  Maughan}.} \bibinfo{year}{2012}\natexlab{}.
\newblock \bibinfo{booktitle}{\emph{{The Menlo Report: Ethical Principles
  Guiding Information and Communication Technology Research}}}.
\newblock \bibinfo{type}{{T}echnical {R}eport}. \bibinfo{institution}{U.S.
  Department of Homeland Security}.
\newblock


\bibitem[Bender et~al\mbox{.}(2021)]%
        {bender2021dangers}
\bibfield{author}{\bibinfo{person}{Emily~M. Bender}, \bibinfo{person}{Timnit
  Gebru}, \bibinfo{person}{Angelina McMillan-Major}, {and}
  \bibinfo{person}{Shmargaret Shmitchell}.} \bibinfo{year}{2021}\natexlab{}.
\newblock \showarticletitle{On the Dangers of Stochastic Parrots: Can Language
  Models Be Too Big?}
\newblock \bibinfo{journal}{\emph{Proceedings of FAccT}}
  (\bibinfo{year}{2021}).
\newblock


\bibitem[Bengio and Raji(2021)]%
        {bengio2021retrospective}
\bibfield{author}{\bibinfo{person}{Samy Bengio} {and} \bibinfo{person}{Deborah
  Raji}.} \bibinfo{year}{2021}\natexlab{}.
\newblock \bibinfo{title}{A Retrospective on the {NeurIPS} 2021 Ethics Review
  Process}.
\newblock
\newblock
\urldef\tempurl%
\url{https://blog.neurips.cc/2021/12/03/a-retrospective-on-the-neurips-2021-ethics-review-process/}
\showURL{%
\tempurl}


\bibitem[Bengtsson(2016)]%
        {bengtsson.2016}
\bibfield{author}{\bibinfo{person}{Mariette Bengtsson}.}
  \bibinfo{year}{2016}\natexlab{}.
\newblock \showarticletitle{How to plan and perform a qualitative study using
  content analysis}.
\newblock \bibinfo{journal}{\emph{NursingPlus Open}}  \bibinfo{volume}{2}
  (\bibinfo{year}{2016}), \bibinfo{pages}{8--14}.
\newblock


\bibitem[Benjamin(2019)]%
        {benjamin2019race}
\bibfield{author}{\bibinfo{person}{Ruha Benjamin}.}
  \bibinfo{year}{2019}\natexlab{}.
\newblock \bibinfo{booktitle}{\emph{Race After Technology: Abolitionist Tools
  for the New Jim Code}}.
\newblock \bibinfo{publisher}{Wiley}.
\newblock


\bibitem[Berg and Lune(2017)]%
        {berg2017qualitative}
\bibfield{author}{\bibinfo{person}{Bruce~L. Berg} {and} \bibinfo{person}{Howard
  Lune}.} \bibinfo{year}{2017}\natexlab{}.
\newblock \bibinfo{booktitle}{\emph{Qualitative research methods for the social
  sciences} (\bibinfo{edition}{ninth edition} ed.)}.
\newblock \bibinfo{publisher}{Pearson}.
\newblock
\showISBNx{9780134202136}


\bibitem[Birhane(2020)]%
        {birhane2020algorithmic}
\bibfield{author}{\bibinfo{person}{Abeba Birhane}.}
  \bibinfo{year}{2020}\natexlab{}.
\newblock \showarticletitle{Algorithmic Colonization of {Africa}}.
\newblock \bibinfo{journal}{\emph{SCRIPTed}} \bibinfo{volume}{17},
  \bibinfo{number}{2} (\bibinfo{year}{2020}).
\newblock


\bibitem[Blodgett et~al\mbox{.}(2020)]%
        {blodgett2020language}
\bibfield{author}{\bibinfo{person}{Su~Lin Blodgett}, \bibinfo{person}{Solon
  Barocas}, \bibinfo{person}{Hal Daum{\'e}~III}, {and} \bibinfo{person}{Hanna
  Wallach}.} \bibinfo{year}{2020}\natexlab{}.
\newblock \showarticletitle{Language (Technology) is Power: {A} Critical Survey
  of {``}Bias{''} in {NLP}}. In \bibinfo{booktitle}{\emph{Proceedings of the
  58th Annual Meeting of the Association for Computational Linguistics}}.
\newblock
\urldef\tempurl%
\url{https://doi.org/10.18653/v1/2020.acl-main.485}
\showDOI{\tempurl}


\bibitem[Bowker and Star(2000)]%
        {bowker2000sorting}
\bibfield{author}{\bibinfo{person}{Geoffrey~C. Bowker} {and}
  \bibinfo{person}{Susan~Leigh Star}.} \bibinfo{year}{2000}\natexlab{}.
\newblock \bibinfo{booktitle}{\emph{Sorting things out: Classification and its
  consequences}}.
\newblock \bibinfo{publisher}{MIT press}.
\newblock


\bibitem[Buolamwini and Gebru(2018)]%
        {buolamwini2018gender}
\bibfield{author}{\bibinfo{person}{Joy Buolamwini} {and}
  \bibinfo{person}{Timnit Gebru}.} \bibinfo{year}{2018}\natexlab{}.
\newblock \showarticletitle{Gender Shades: {Intersectional} Accuracy
  Disparities in Commercial Gender Classification}. In
  \bibinfo{booktitle}{\emph{Proceedings of the Conference on Fairness,
  Accountability and Transparency}}.
\newblock


\bibitem[Costanza-Chock(2018)]%
        {costanza2018design}
\bibfield{author}{\bibinfo{person}{Sasha Costanza-Chock}.}
  \bibinfo{year}{2018}\natexlab{}.
\newblock \showarticletitle{Design Justice, {AI}, and Escape from the Matrix of
  Domination}.
\newblock \bibinfo{journal}{\emph{Journal of Design and Science}}
  (\bibinfo{year}{2018}).
\newblock


\bibitem[Crawford(2017)]%
        {crawford.2017}
\bibfield{author}{\bibinfo{person}{Kate Crawford}.}
  \bibinfo{year}{2017}\natexlab{}.
\newblock \bibinfo{title}{The Trouble with Bias}.  (\bibinfo{year}{2017}).
\newblock
\newblock
\shownote{NeurIPS Keynote}.


\bibitem[Denzin(2017)]%
        {denzin2017sociological}
\bibfield{author}{\bibinfo{person}{Norman~K Denzin}.}
  \bibinfo{year}{2017}\natexlab{}.
\newblock \bibinfo{booktitle}{\emph{Sociological methods: a sourcebook}}.
\newblock \bibinfo{publisher}{McGraw-Hill}.
\newblock


\bibitem[D'Ignazio and Klein(2020)]%
        {d2020data}
\bibfield{author}{\bibinfo{person}{Catherine D'Ignazio} {and}
  \bibinfo{person}{Lauren~F Klein}.} \bibinfo{year}{2020}\natexlab{}.
\newblock \bibinfo{booktitle}{\emph{Data Feminism}}.
\newblock \bibinfo{publisher}{MIT Press}.
\newblock


\bibitem[Dotan and Milli(2019)]%
        {dotan2019value}
\bibfield{author}{\bibinfo{person}{Ravit Dotan} {and} \bibinfo{person}{Smitha
  Milli}.} \bibinfo{year}{2019}\natexlab{}.
\newblock \showarticletitle{Value-Laden Disciplinary Shifts in Machine
  Learning}.
\newblock \bibinfo{journal}{\emph{arXiv preprint arXiv:1912.01172}}
  (\bibinfo{year}{2019}).
\newblock


\bibitem[Doyle et~al\mbox{.}(2020)]%
        {doyle2020overview}
\bibfield{author}{\bibinfo{person}{Louise Doyle}, \bibinfo{person}{Catherine
  McCabe}, \bibinfo{person}{Brian Keogh}, \bibinfo{person}{Annemarie Brady},
  {and} \bibinfo{person}{Margaret McCann}.} \bibinfo{year}{2020}\natexlab{}.
\newblock \showarticletitle{An overview of the qualitative descriptive design
  within nursing research}.
\newblock \bibinfo{journal}{\emph{Journal of Research in Nursing}}
  \bibinfo{volume}{25}, \bibinfo{number}{5} (\bibinfo{date}{Aug}
  \bibinfo{year}{2020}), \bibinfo{pages}{443–455}.
\newblock
\showISSN{1744-9871, 1744-988X}
\urldef\tempurl%
\url{https://doi.org/10.1177/1744987119880234}
\showDOI{\tempurl}


\bibitem[Ethayarajh and Jurafsky(2020)]%
        {ethayarajh.2020}
\bibfield{author}{\bibinfo{person}{Kawin Ethayarajh} {and} \bibinfo{person}{Dan
  Jurafsky}.} \bibinfo{year}{2020}\natexlab{}.
\newblock \showarticletitle{Utility is in the Eye of the User: {A} Critique of
  {NLP} Leaderboards}. In \bibinfo{booktitle}{\emph{Proceedings of EMNLP}}.
\newblock


\bibitem[Floridi and Cowls(2019)]%
        {floridi.2019}
\bibfield{author}{\bibinfo{person}{Luciano Floridi} {and} \bibinfo{person}{Josh
  Cowls}.} \bibinfo{year}{2019}\natexlab{}.
\newblock \showarticletitle{A Unified Framework of Five Principles for {AI} in
  Society}.
\newblock \bibinfo{journal}{\emph{Harvard Data Science Review}}
  \bibinfo{volume}{1}, \bibinfo{number}{1} (\bibinfo{year}{2019}).
\newblock


\bibitem[Glaser and Strauss(1999)]%
        {glaser1999discovery}
\bibfield{author}{\bibinfo{person}{Barney~G. Glaser} {and}
  \bibinfo{person}{Anselm~L. Strauss}.} \bibinfo{year}{1999}\natexlab{}.
\newblock \bibinfo{booktitle}{\emph{The discovery of grounded theory:
  strategies for grounded research}}.
\newblock \bibinfo{publisher}{Aldine de Gruyter}.
\newblock


\bibitem[Green(2019)]%
        {green2019good}
\bibfield{author}{\bibinfo{person}{Ben Green}.}
  \bibinfo{year}{2019}\natexlab{}.
\newblock \showarticletitle{{`Good'} isn’t Good Enough}. In
  \bibinfo{booktitle}{\emph{NeurIPS Joint Workshop on AI for Social Good}}.
\newblock


\bibitem[Hamidi et~al\mbox{.}(2018)]%
        {hamidi2018gender}
\bibfield{author}{\bibinfo{person}{Foad Hamidi}, \bibinfo{person}{Morgan~Klaus
  Scheuerman}, {and} \bibinfo{person}{Stacy~M. Branham}.}
  \bibinfo{year}{2018}\natexlab{}.
\newblock \showarticletitle{Gender Recognition or Gender Reductionism? {T}he
  Social Implications of Embedded Gender Recognition Systems}. In
  \bibinfo{booktitle}{\emph{Proceedings CHI}}.
\newblock


\bibitem[Hanna and Park(2020)]%
        {hanna2020against}
\bibfield{author}{\bibinfo{person}{Alex Hanna} {and} \bibinfo{person}{Tina~M.
  Park}.} \bibinfo{year}{2020}\natexlab{}.
\newblock \showarticletitle{Against Scale: Provocations and Resistances to
  Scale Thinking}.
\newblock \bibinfo{journal}{\emph{arXiv preprint arXiv:2010.08850}}
  (\bibinfo{year}{2020}).
\newblock


\bibitem[Hill(2020)]%
        {hill2020accused}
\bibfield{author}{\bibinfo{person}{Kashmir Hill}.}
  \bibinfo{year}{2020}\natexlab{}.
\newblock \showarticletitle{Wrongfully Accused by an Algorithm}.
\newblock \bibinfo{journal}{\emph{The New York Times}} (\bibinfo{year}{2020}).
\newblock
\urldef\tempurl%
\url{https://www.nytimes.com/2020/06/24/technology/facial-recognition-arrest.html}
\showURL{%
\tempurl}


\bibitem[Holstein et~al\mbox{.}(2019)]%
        {holstein.2019}
\bibfield{author}{\bibinfo{person}{Kenneth Holstein}, \bibinfo{person}{Jennifer
  Wortman~Vaughan}, \bibinfo{person}{Hal Daum\'{e}~III}, \bibinfo{person}{Miro
  Dudik}, {and} \bibinfo{person}{Hanna Wallach}.}
  \bibinfo{year}{2019}\natexlab{}.
\newblock \showarticletitle{Improving Fairness in Machine Learning Systems:
  What Do Industry Practitioners Need?}. In
  \bibinfo{booktitle}{\emph{Proceedings of CHI}}.
\newblock


\bibitem[Hsieh and Shannon(2005)]%
        {hsieh2005three}
\bibfield{author}{\bibinfo{person}{Hsiu-Fang Hsieh} {and}
  \bibinfo{person}{Sarah~E. Shannon}.} \bibinfo{year}{2005}\natexlab{}.
\newblock \showarticletitle{Three Approaches to Qualitative Content Analysis}.
\newblock \bibinfo{journal}{\emph{Qualitative Health Research}}
  \bibinfo{volume}{15}, \bibinfo{number}{9} (\bibinfo{year}{2005}),
  \bibinfo{pages}{1277–1288}.
\newblock
\urldef\tempurl%
\url{https://doi.org/10.1177/1049732305276687}
\showDOI{\tempurl}


\bibitem[Ioannidis(2012)]%
        {ioannidis.2012}
\bibfield{author}{\bibinfo{person}{John P.~A. Ioannidis}.}
  \bibinfo{year}{2012}\natexlab{}.
\newblock \showarticletitle{Why Science Is Not Necessarily Self-Correcting}.
\newblock \bibinfo{journal}{\emph{Perspectives on Psychological Science}}
  \bibinfo{volume}{7}, \bibinfo{number}{6} (\bibinfo{year}{2012}),
  \bibinfo{pages}{645--654}.
\newblock


\bibitem[Kalluri(2019)]%
        {kalluri2019values}
\bibfield{author}{\bibinfo{person}{Pratyusha Kalluri}.}
  \bibinfo{year}{2019}\natexlab{}.
\newblock \bibinfo{title}{The Values of Machine Learning}.
\newblock
\newblock
\urldef\tempurl%
\url{https://slideslive.com/38923453/the-values-of-machine-learning}
\showURL{%
\tempurl}
\newblock
\shownote{https://slideslive.com/38923453/the-values-of-machine-learning}.


\bibitem[Kalluri(2020)]%
        {kalluri2020don}
\bibfield{author}{\bibinfo{person}{Pratyusha Kalluri}.}
  \bibinfo{year}{2020}\natexlab{}.
\newblock \showarticletitle{Don't ask if Artificial Intelligence is Good or
  Fair, ask how it Shifts Power.}
\newblock \bibinfo{journal}{\emph{Nature}} \bibinfo{volume}{583},
  \bibinfo{number}{7815} (\bibinfo{year}{2020}), \bibinfo{pages}{169--169}.
\newblock


\bibitem[Kendall and Smith(1939)]%
        {kendall1939problem}
\bibfield{author}{\bibinfo{person}{Maurice~G Kendall} {and}
  \bibinfo{person}{B~Babington Smith}.} \bibinfo{year}{1939}\natexlab{}.
\newblock \showarticletitle{The problem of m rankings}.
\newblock \bibinfo{journal}{\emph{The annals of mathematical statistics}}
  \bibinfo{volume}{10}, \bibinfo{number}{3} (\bibinfo{year}{1939}),
  \bibinfo{pages}{275--287}.
\newblock


\bibitem[Kirilenko and Stepchenkova(2016)]%
        {kirilenko2016inter}
\bibfield{author}{\bibinfo{person}{Andrei~P Kirilenko} {and}
  \bibinfo{person}{Svetlana Stepchenkova}.} \bibinfo{year}{2016}\natexlab{}.
\newblock \showarticletitle{Inter-coder agreement in one-to-many
  classification: fuzzy kappa}.
\newblock \bibinfo{journal}{\emph{PloS one}} \bibinfo{volume}{11},
  \bibinfo{number}{3} (\bibinfo{year}{2016}), \bibinfo{pages}{e0149787}.
\newblock


\bibitem[Krefting(1991)]%
        {krefting.1991}
\bibfield{author}{\bibinfo{person}{Laura Krefting}.}
  \bibinfo{year}{1991}\natexlab{}.
\newblock \showarticletitle{Rigor in Qualitative Research: {T}he Assessment of
  Trustworthiness}.
\newblock \bibinfo{journal}{\emph{American Journal of Occupational Therapy}}
  \bibinfo{volume}{45}, \bibinfo{number}{3} (\bibinfo{date}{03}
  \bibinfo{year}{1991}), \bibinfo{pages}{214--222}.
\newblock
\urldef\tempurl%
\url{https://doi.org/10.5014/ajot.45.3.214}
\showDOI{\tempurl}


\bibitem[Krippendorff(2018)]%
        {krippendorff.2018}
\bibfield{author}{\bibinfo{person}{Klaus Krippendorff}.}
  \bibinfo{year}{2018}\natexlab{}.
\newblock \bibinfo{booktitle}{\emph{Content Analysis: {An} Introduction to its
  Methodology}}.
\newblock \bibinfo{publisher}{Sage Publications}.
\newblock


\bibitem[Kuhn(1977)]%
        {kuhn.1977}
\bibfield{author}{\bibinfo{person}{Thomas~S. Kuhn}.}
  \bibinfo{year}{1977}\natexlab{}.
\newblock \showarticletitle{Objectivity, Value Judgment, and Theory Choice}.
\newblock In \bibinfo{booktitle}{\emph{The Essential Tension: Selected Studies
  in Scientific Tradition and Change}}. \bibinfo{publisher}{University of
  Chicago Press}, \bibinfo{pages}{320--39}.
\newblock


\bibitem[Lewis et~al\mbox{.}(2020)]%
        {lewis2020indigenous}
\bibfield{author}{\bibinfo{person}{Jason~Edward Lewis}, \bibinfo{person}{Angie
  Abdilla}, \bibinfo{person}{Noelani Arista},
  \bibinfo{person}{Kaipulaumakaniolono Baker}, \bibinfo{person}{Scott
  Benesiinaabandan}, \bibinfo{person}{Michelle Brown}, \bibinfo{person}{Melanie
  Cheung}, \bibinfo{person}{Meredith Coleman}, \bibinfo{person}{Ashley Cordes},
  \bibinfo{person}{Joel Davison}, {et~al\mbox{.}}}
  \bibinfo{year}{2020}\natexlab{}.
\newblock \showarticletitle{Indigenous Protocol and Artificial Intelligence
  Position Paper}.
\newblock  (\bibinfo{year}{2020}).
\newblock


\bibitem[Lewis et~al\mbox{.}(2018)]%
        {lewis2018digital}
\bibfield{author}{\bibinfo{person}{T. Lewis}, \bibinfo{person}{S.~P.
  Gangadharan}, \bibinfo{person}{M. Saba}, {and} \bibinfo{person}{T. Petty}.}
  \bibinfo{year}{2018}\natexlab{}.
\newblock \bibinfo{booktitle}{\emph{Digital Defense Playbook: Community power
  tools for reclaiming data}}.
\newblock \bibinfo{publisher}{Our Data Bodies}.
\newblock


\bibitem[Lincoln and Guba(2006)]%
        {lincoln2006naturalistic}
\bibfield{author}{\bibinfo{person}{Yvonna~S. Lincoln} {and}
  \bibinfo{person}{Egon~G. Guba}.} \bibinfo{year}{2006}\natexlab{}.
\newblock \bibinfo{booktitle}{\emph{Naturalistic inquiry}}.
\newblock \bibinfo{publisher}{Sage Publ.}
\newblock


\bibitem[Lipton and Steinhardt(2019)]%
        {lipton.2018}
\bibfield{author}{\bibinfo{person}{Zachary~C. Lipton} {and}
  \bibinfo{person}{Jacob Steinhardt}.} \bibinfo{year}{2019}\natexlab{}.
\newblock \showarticletitle{Troubling Trends in Machine Learning Scholarship:
  {S}ome {ML} Papers Suffer from Flaws That Could Mislead the Public and Stymie
  Future Research}.
\newblock \bibinfo{journal}{\emph{Queue}} \bibinfo{volume}{17},
  \bibinfo{number}{1} (\bibinfo{year}{2019}), \bibinfo{pages}{45–77}.
\newblock
\urldef\tempurl%
\url{https://doi.org/10.1145/3317287.3328534}
\showDOI{\tempurl}


\bibitem[Longino(1996)]%
        {longino.1996}
\bibfield{author}{\bibinfo{person}{Helen~E. Longino}.}
  \bibinfo{year}{1996}\natexlab{}.
\newblock \showarticletitle{Cognitive and Non-Cognitive Values in Science:
  Rethinking the Dichotomy}.
\newblock In \bibinfo{booktitle}{\emph{Feminism, Science, and the Philosophy of
  Science}}, \bibfield{editor}{\bibinfo{person}{Lynn~Hankinson Nelson} {and}
  \bibinfo{person}{Jack Nelson}} (Eds.). \bibinfo{publisher}{Springer
  Netherlands}, \bibinfo{pages}{39--58}.
\newblock


\bibitem[McMullin(1982)]%
        {mcmullin1982values}
\bibfield{author}{\bibinfo{person}{Ernan McMullin}.}
  \bibinfo{year}{1982}\natexlab{}.
\newblock \showarticletitle{Values in science}. In
  \bibinfo{booktitle}{\emph{Proceedings of the Biennial Meeting of the
  Philosophy of Science Association}}.
\newblock


\bibitem[Merriam and Grenier(2019)]%
        {merriam2019qualitative}
\bibfield{author}{\bibinfo{person}{Sharan~B Merriam} {and}
  \bibinfo{person}{Robin~S Grenier}.} \bibinfo{year}{2019}\natexlab{}.
\newblock \bibinfo{booktitle}{\emph{Qualitative Research in Practice}}.
\newblock \bibinfo{publisher}{Jossey-Bass}.
\newblock


\bibitem[Merton(1973)]%
        {merton.1973}
\bibfield{author}{\bibinfo{person}{Robert~K. Merton}.}
  \bibinfo{year}{1973}\natexlab{}.
\newblock \bibinfo{booktitle}{\emph{The Sociology of Science: Theoretical and
  Empirical Investigations}}.
\newblock \bibinfo{publisher}{University of Chicago press}.
\newblock


\bibitem[Mohamed et~al\mbox{.}(2020)]%
        {mohamed2020decolonial}
\bibfield{author}{\bibinfo{person}{Shakir Mohamed},
  \bibinfo{person}{Marie-Therese Png}, {and} \bibinfo{person}{William Isaac}.}
  \bibinfo{year}{2020}\natexlab{}.
\newblock \showarticletitle{Decolonial {AI}: {Decolonial} Rheory as
  Sociotechnical Foresight in Artificial Intelligence}.
\newblock \bibinfo{journal}{\emph{Philosophy \& Technology}}
  \bibinfo{volume}{33} (\bibinfo{year}{2020}), \bibinfo{pages}{659--–684}.
\newblock


\bibitem[Nanayakkara et~al\mbox{.}(2021)]%
        {nanayakkara.2021}
\bibfield{author}{\bibinfo{person}{Priyanka Nanayakkara},
  \bibinfo{person}{Jessica Hullman}, {and} \bibinfo{person}{Nicholas
  Diakopoulos}.} \bibinfo{year}{2021}\natexlab{}.
\newblock \showarticletitle{Unpacking the Expressed Consequences of {AI}
  Research in Broader Impact Statements}. In
  \bibinfo{booktitle}{\emph{Proceedings of the 2021 AAAI/ACM Conference on AI,
  Ethics, and Society}}.
\newblock
\urldef\tempurl%
\url{https://doi.org/10.1145/3461702.3462608}
\showURL{%
\tempurl}


\bibitem[Noble and Smith(2015)]%
        {noble2015issues}
\bibfield{author}{\bibinfo{person}{Helen Noble} {and} \bibinfo{person}{Joanna
  Smith}.} \bibinfo{year}{2015}\natexlab{}.
\newblock \showarticletitle{Issues of validity and reliability in qualitative
  research}.
\newblock \bibinfo{journal}{\emph{Evidence Based Nursing}}
  \bibinfo{volume}{18}, \bibinfo{number}{2} (\bibinfo{date}{Apr}
  \bibinfo{year}{2015}), \bibinfo{pages}{34–35}.
\newblock


\bibitem[Noble(2018)]%
        {noble2018algorithms}
\bibfield{author}{\bibinfo{person}{Safiya~Umoja Noble}.}
  \bibinfo{year}{2018}\natexlab{}.
\newblock \bibinfo{booktitle}{\emph{Algorithms of oppression: How search
  engines reinforce racism}}.
\newblock \bibinfo{publisher}{{NYU Press}}.
\newblock


\bibitem[O'Neil(2016)]%
        {o2016weapons}
\bibfield{author}{\bibinfo{person}{Cathy O'Neil}.}
  \bibinfo{year}{2016}\natexlab{}.
\newblock \bibinfo{booktitle}{\emph{Weapons of Math Destruction: How Big Data
  Increases Inequality and Threatens Democracy}}.
\newblock \bibinfo{publisher}{Broadway Books}.
\newblock


\bibitem[Pasquale(2015)]%
        {pasquale2015black}
\bibfield{author}{\bibinfo{person}{Frank Pasquale}.}
  \bibinfo{year}{2015}\natexlab{}.
\newblock \bibinfo{booktitle}{\emph{The black box society}}.
\newblock \bibinfo{publisher}{Harvard University Press}.
\newblock


\bibitem[Patton(1990)]%
        {patton1990qualitative}
\bibfield{author}{\bibinfo{person}{Michael~Quinn Patton}.}
  \bibinfo{year}{1990}\natexlab{}.
\newblock \bibinfo{booktitle}{\emph{Qualitative Evaluation and Research
  Methods}}.
\newblock \bibinfo{publisher}{Sage}.
\newblock


\bibitem[Patton(1999)]%
        {patton1999enhancing}
\bibfield{author}{\bibinfo{person}{M~Q Patton}.}
  \bibinfo{year}{1999}\natexlab{}.
\newblock \showarticletitle{Enhancing the quality and credibility of
  qualitative analysis}.
\newblock \bibinfo{journal}{\emph{Health Services Research}}
  \bibinfo{volume}{34}, \bibinfo{number}{5} (\bibinfo{date}{Dec}
  \bibinfo{year}{1999}).
\newblock


\bibitem[Perdomo et~al\mbox{.}(2020)]%
        {perdomo2020performative}
\bibfield{author}{\bibinfo{person}{Juan Perdomo}, \bibinfo{person}{Tijana
  Zrnic}, \bibinfo{person}{Celestine Mendler-D{\"u}nner}, {and}
  \bibinfo{person}{Moritz Hardt}.} \bibinfo{year}{2020}\natexlab{}.
\newblock \showarticletitle{Performative Prediction}. In
  \bibinfo{booktitle}{\emph{Proceedings of ICML}}.
\newblock


\bibitem[Prabhu and Birhane(2020)]%
        {prabhu2020large}
\bibfield{author}{\bibinfo{person}{Vinay~Uday Prabhu} {and}
  \bibinfo{person}{Abeba Birhane}.} \bibinfo{year}{2020}\natexlab{}.
\newblock \showarticletitle{Large Image Datasets: {A} {P}yrrhic Win for
  Computer Vision?}
\newblock \bibinfo{journal}{\emph{arXiv preprint arXiv:2006.16923}}
  (\bibinfo{year}{2020}).
\newblock
\urldef\tempurl%
\url{https://arxiv.org/abs/2006.16923}
\showURL{%
\tempurl}


\bibitem[Rogaway(2015)]%
        {rogaway2015moral}
\bibfield{author}{\bibinfo{person}{Phillip Rogaway}.}
  \bibinfo{year}{2015}\natexlab{}.
\newblock \bibinfo{title}{The Moral Character of Cryptographic Work}.
\newblock \bibinfo{howpublished}{Cryptology ePrint Archive, Report 2015/1162}.
\newblock
\newblock
\shownote{\url{https://eprint.iacr.org/2015/1162}}.


\bibitem[Rogers(2019)]%
        {rogers.2020}
\bibfield{author}{\bibinfo{person}{Anna Rogers}.}
  \bibinfo{year}{2019}\natexlab{}.
\newblock \showarticletitle{Peer review in {NLP}: reject-if-not-{SOTA}}.
\newblock \bibinfo{journal}{\emph{Hacking Semantics blog}}
  (\bibinfo{year}{2019}).
\newblock
\urldef\tempurl%
\url{https://hackingsemantics.xyz/2020/reviewing-models/#everything-wrong-with-reject-if-not-sota}
\showURL{%
\tempurl}


\bibitem[Rus(2018)]%
        {rus.2018}
\bibfield{author}{\bibinfo{person}{Daniela Rus}.}
  \bibinfo{year}{2018}\natexlab{}.
\newblock \showarticletitle{Rise of the robots: {A}re you ready?}
\newblock \bibinfo{journal}{\emph{Financial Times Magazine}}
  (\bibinfo{date}{March} \bibinfo{year}{2018}).
\newblock
\urldef\tempurl%
\url{https://www.ft.com/content/e31c4986-20d0-11e8-a895-1ba1f72c2c11}
\showURL{%
\tempurl}


\bibitem[Suresh and Guttag(2019)]%
        {suresh2019framework}
\bibfield{author}{\bibinfo{person}{Harini Suresh} {and}
  \bibinfo{person}{John~V. Guttag}.} \bibinfo{year}{2019}\natexlab{}.
\newblock \showarticletitle{A Framework for Understanding Unintended
  Consequences of Machine Learning}.
\newblock \bibinfo{journal}{\emph{arXiv preprint arXiv:1901.10002}}
  (\bibinfo{year}{2019}).
\newblock
\urldef\tempurl%
\url{http://arxiv.org/abs/1901.10002}
\showURL{%
\tempurl}


\bibitem[Trapido(2015)]%
        {trapido.2015}
\bibfield{author}{\bibinfo{person}{Denis Trapido}.}
  \bibinfo{year}{2015}\natexlab{}.
\newblock \showarticletitle{How Novelty in Knowledge Earns Recognition: {The}
  Role of Consistent Identities}.
\newblock \bibinfo{journal}{\emph{Research Policy}} \bibinfo{volume}{44},
  \bibinfo{number}{8} (\bibinfo{year}{2015}), \bibinfo{pages}{1488--1500}.
\newblock


\bibitem[Vinkers et~al\mbox{.}(2015)]%
        {vinkers.2015}
\bibfield{author}{\bibinfo{person}{Christiaan~H. Vinkers},
  \bibinfo{person}{Joeri~K. Tijdink}, {and} \bibinfo{person}{Willem~M. Otte}.}
  \bibinfo{year}{2015}\natexlab{}.
\newblock \showarticletitle{Use of Positive and Negative Words in Scientific
  {PubMed} Abstracts between 1974 and 2014: Retrospective Analysis}.
\newblock \bibinfo{journal}{\emph{BMJ}}  \bibinfo{volume}{351}
  (\bibinfo{year}{2015}).
\newblock


\bibitem[Wagstaff(2012)]%
        {wagstaff.2012}
\bibfield{author}{\bibinfo{person}{Kiri Wagstaff}.}
  \bibinfo{year}{2012}\natexlab{}.
\newblock \showarticletitle{Machine Learning that Matters}. In
  \bibinfo{booktitle}{\emph{Proceedings of ICML}}.
\newblock


\bibitem[Weizenbaum(1972)]%
        {weizenbaum1972impact}
\bibfield{author}{\bibinfo{person}{Joseph Weizenbaum}.}
  \bibinfo{year}{1972}\natexlab{}.
\newblock \showarticletitle{On the Impact of the Computer on Society}.
\newblock \bibinfo{journal}{\emph{Science}} \bibinfo{volume}{176},
  \bibinfo{number}{4035} (\bibinfo{year}{1972}), \bibinfo{pages}{609--614}.
\newblock


\bibitem[Winner(1977)]%
        {winner.1977}
\bibfield{author}{\bibinfo{person}{Langdon Winner}.}
  \bibinfo{year}{1977}\natexlab{}.
\newblock \bibinfo{booktitle}{\emph{Autonomous Technology:
  Technics-out-of-Control as a Theme in Political Thought}}.
\newblock \bibinfo{publisher}{{MIT} Press}.
\newblock


\bibitem[Winner(1980)]%
        {winner1980artifacts}
\bibfield{author}{\bibinfo{person}{Langdon Winner}.}
  \bibinfo{year}{1980}\natexlab{}.
\newblock \showarticletitle{Do Artifacts Have Politics?}
\newblock \bibinfo{journal}{\emph{Daedalus}} \bibinfo{volume}{109},
  \bibinfo{number}{1} (\bibinfo{year}{1980}), \bibinfo{pages}{121--136}.
\newblock


\end{thebibliography}

\renewcommand*{\thesection}{\Alph{section}}

\newpage
\appendix
\renewcommand\thefigure{\thesection.\arabic{figure}}


\setcounter{figure}{0}

\section{Additional Methodological Details}

\subsection{Data Sources}
\label{sec:data}

To determine the most-cited papers from each conference, we rely on the publicly-available Semantic Scholar database \citep{ammar.2018}, which includes bibliographic information for scientific papers, including citation counts.\footnote{\url{http://s2-public-api.prod.s2.allenai.org/corpus/}}
Using this data, we chose the most cited papers from each of 2008, 2009, 2018, 2019 published at NeurIPS and ICML. 

Like all bibliographic databases, Semantic Scholar is imperfect. Upon manual review, we wish to document that our selection includes one paper that was actually published in 2010, and one that was retracted from NeurIPS prior to publication (see \S\ref{sec:paperlist} for details). In addition, the citations counts used to determine the most cited papers reflect a static moment in time, and may differ from other sources. 

Because our artifacts of study are papers that have been previously published at NeurIPS or ICML, we surmise that the authors normatively expect and consent to their papers and themselves as authors being referenced and analyzed in future papers, e.g. this paper. Accordingly, we chose not to seek explicit permission from the original authors to reference, annotate, and analyze their papers. The annotations we generated do not introduce any new personally identifying information or offensive content. The sentences from the original published papers are necessarily part of our annotations; to the extent that these papers have these issues, these sentences may contain personally identifying information or offensive content. Given the original authors contributed their work to the same venues as our own work, we believe that the potential to cause new harm from this inclusion is minimal.

\subsection{Defining elite university}
To determine the list of elite universities, we follow Ahmed and Wahed \citep{ahmed2020democratization}, and rely on the QS World University Rankings for the discipline of computer science. For 2018/19, we take the top 50 schools from the CS rankings for 2018. For 2008/09, we take the top 50 schools from the CS rankings for 2011, as the closest year for which data is available. 

\subsection{Defining big tech}
We used Abdalla and Abdalla's \citep{abdalla2020grey} criterion to what is considered ``big tech'', which is comprised of: Alibaba, Amazon, Apple, Element AI, Facebook, Google, Huawei, IBM, Intel, Microsoft, Nvidia, Open AI, Samsung, and Uber. Furthermore, we added DeepMind to this list, which Google acquired in 2014. We considered all other companies as ``non-big tech''.

\section{Annotations}
We include the annotations of all papers 
as supplementary material at \url{https://github.com/wagnew3/The-Values-Encoded-in-Machine-Learning-Research} with a CC BY-NC-SA license. 
To present a birds-eye view of the value annotations, we present randomly selected examples of annotated sentences in Appendix \ref{sec:randomexamples}. 

\section{Combining values}
\label{sec:value_clusters}
\begin{figure*}[h!]
\centering
\includegraphics[origin=c,width=0.72\linewidth]{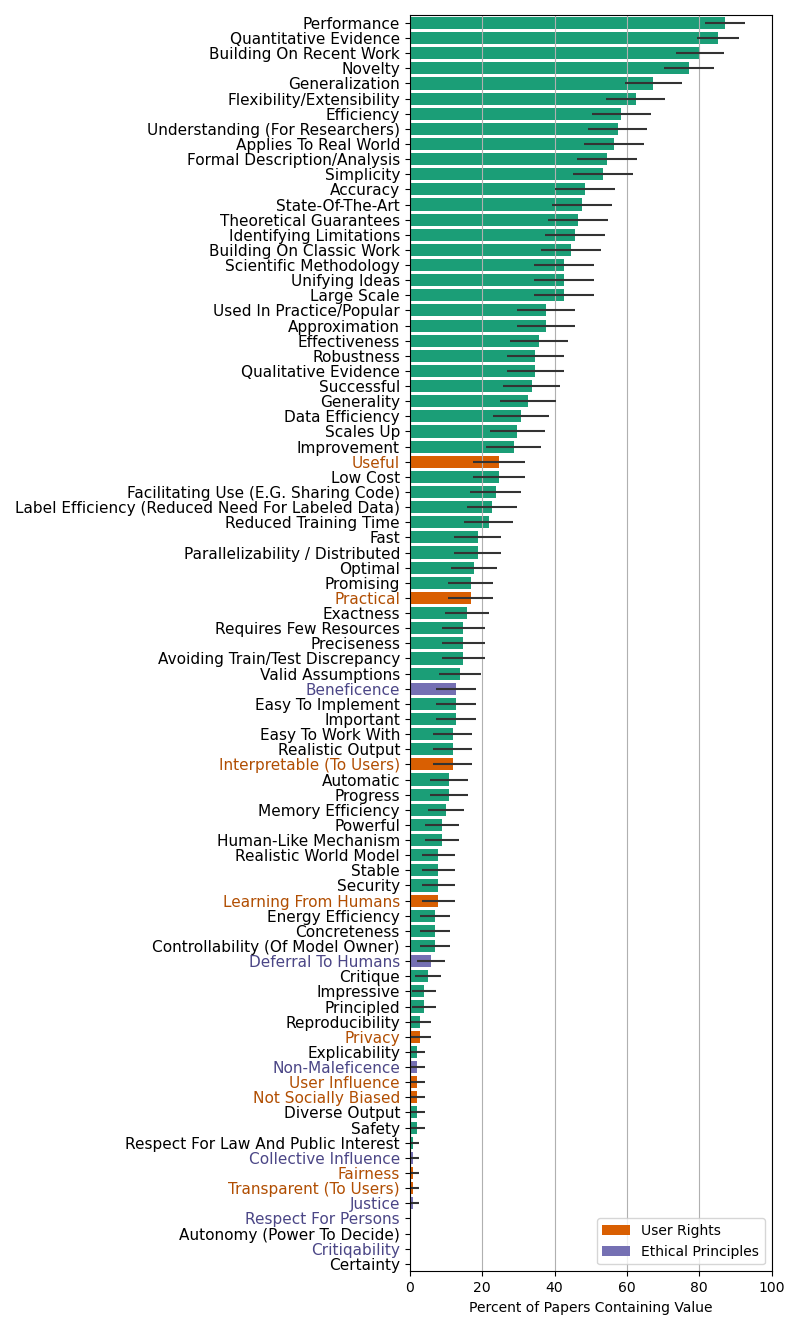}
\caption{Proportion of annotated papers that uplifted each value, prior to combining.}
\label{fig:value-totals-all}
\end{figure*}

In some cases, 
values had strong overlap with related values (e.g., \textit{Performance} is closely related to  \textit{Accuracy}). In other cases, 
we had annotated for several fine-grained values that we found could be combined (e.g., \emph{Data Efficiency} and \textit{Label Efficiency} are types of \emph{Efficiency}). Following best practices, we identified such values and combined them for our main analysis. In Figure~\ref{fig:value-totals-all}
 we list all values before combining, and in Table~\ref{tab:value_clusters} we list the combinations.

\begin{table*}[h]
    \centering
    \small
    \caption{Sets of values combined for analysis in the main paper}
    \begin{tabular}{l l}
        Overarching value & Set of values  \\
        \hline
        Performance & Performance, Accuracy, State-of-the-art \\
        Building on past work & Building on classic work, Building on recent work \\
        Generalization & Generalization, Avoiding train/test discrepancy, Flexibility/extensibility  \\
        Efficiency & Efficiency, Data efficiency, Energy efficiency, Fast, Label efficiency, \\
        & Low cost, Memory efficiency, Reduced training time \\
    \end{tabular}
    \label{tab:value_clusters}
\end{table*}

\section{Reflexivity statement}
\label{app:reflexivity}
The cloak of objectivity is an important part of what we are challenging in this paper. We are encouraging all researchers to reflect on what norms, perspectives, privileges, or incentives may be shaping their work. By sharing a bit about ourselves, we help make it possible for others to find biases we might not have recognized, and we hope to create space for other peoples and values at the margins. Our team is multi-racial and multi-gender and includes undergraduate, graduate, and post-graduate researchers engaged with AI, machine learning, NLP, robotics, cognitive science, critical theory, grassroots community organizing, abolitionist community organizing, arts, and philosophy of science. We are privileged due to our affiliations with well resourced Western universities enabling our research and its eventual publication with relative ease (for example, compared to the challenges our peers in the global South might have faced). Furthermore, as these Western universities have a history of racism, colonialism and white-supremacy, being embedded in such ecology makes it impossible to entirely sever our research from such histories. Notably for our work, it is possible that different authors might have identified a somewhat different set of values, and/or recognized these values differently in the text, and we acknowledge that we may have gaps in understanding with respect to what is important to others, most importantly to us, what is important to our and other communities at the margins. 
In addition, we recognize that the notion of an ``elite'' university, which we have adopted from past work (both the category and the members), implies a ranking or hierarchy among institutions of higher education that may be unjust to many others. 
Throughout the paper, we have attempted to adopt a critical perspective, but it is likely that there are still many parts of the broader machine learning ecosystem that we simply take for granted, and which should be recognized and challenged.

\section{Experiments with Using Text Classification to Identify Values}
\label{app:automatic}

Although it was not our primary purpose in annotating highly-cited papers, we include here a brief report on using the annotations we generated as potential training data for classifiers that could in principle be used to estimate the prevalence of these values in a larger set of ML papers. This is something that we should approach with great caution for several reasons: i) we only have a relatively small training set of annotated examples with respect to machine learning best practices;
ii) these annotations are taken from a non-random set of papers, and any models trained on these data may not generalize to all papers; iii) an automated approach will fail to detect additional, previously unobserved, emergent values; and iv) based on our experiences annotating these papers, we expect that many would be expressed subtly and in varied ways that would be difficult to detect automatically, at least without considerably more training data.

To present a baseline for testing the potential of this approach, while avoiding any biases that might be introduced by pretrained language models, we make use of simple regularized logistic regression classifiers operating on unigram features. We trained models separately for each value (for all values that had at least 20 relevant sentences, using all relevant sentences for the higher-order grouped values), treating each sentence as an instance with a binary label (present or not), tokenizing each sentence using \emph{spaCy} and converting each to a binary feature representation indicating the presence or absence of each word in the vocabulary (all words occurring at least twice in the corpus). These choices were not tuned. We randomly selected 300 sentences to use as a held out test set (using the same test set for each value), and trained a model using the remaining data, using 5-fold cross validation to tune the regularization strength.

\begin{figure*}[h!]
\centering
\includegraphics[width=1.0\textwidth]{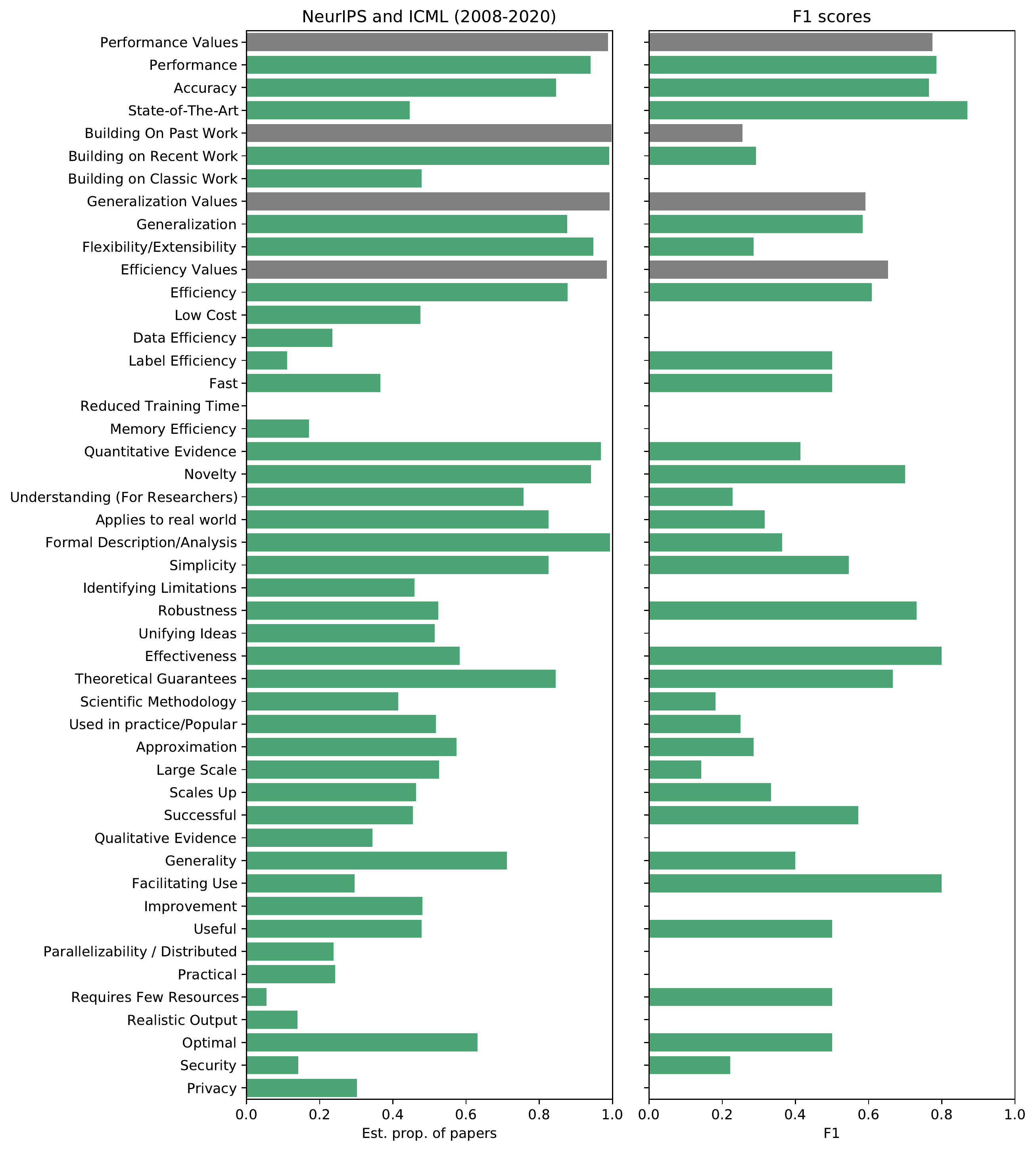}
\caption{Proportion of papers in from 2008--2020 (combining NeurIPS and ICML) predicted to have at least one sentence expressing each value (left), and estimated performance (F1) of the corresponding classifiers (right). Note that the overall performance of most classifiers is generally poor, indicating that the estimates on the left should be treated as unreliable in most cases. Grey bars represent the clustered values. Classifiers were not trained for values with less than 20 representative sentences.}
\label{fig:textcat}
\end{figure*}

F1 scores on the test set for the various models are shown in Figure \ref{fig:textcat} (right), and can generally be seen to be unimpressive. 
The F1 score for most values is on the order of 0.5 or less, and some values, even relatively common ones such as \textit{Unifying Ideas}, ended up with an F1 score of 0. The most highly-weighted features for most classifiers were quite reasonable, but this is evidently a relatively difficult task, at least given this amount of data. The exceptions to this poor performance included the Performance-related values (\emph{Performance}, \emph{Accuracy}, and \emph{State-of-the-art}), as well as \emph{Effectiveness}, and \emph{Facilitating Use}, all of which had F1 scores greater than 0.75, and most of which were typically represented by a relatively small set of terms (e.g., "accurate", "accuracy", "accurately", "inaccurate", "accuracies", "errors", etc. for \emph{Accuracy}).

Although the poor performance of these classifiers means we should interpret any use of them with caution, we explore applying them to a broader set of papers for the sake of completeness. To do so, we download pdfs of all papers published at NeurIPS and ICML for the years 2008 through 2020, convert these to text using \emph{pdftotext}, and extract sentences from this text, excluding references, as well as very short sentences (less than 6 tokens) or lines without alphabetic characters. Note that due to the difficulty of automatically parsing papers into sections, these textual representations are not limited to the abstract, introduction, discussion, and conclusion, in contrast to our annotations, thus we would expect most values to occur more frequently, especially those that are likely to occur in sections about experiments and results. 

We then apply the classifiers trained above to each sentence in each paper. For each value, we then compute the proportion of papers (combining NeurIPS and ICML for this entire time period) that had at least one sentence predicted to exhibit that value. The overall proportions are shown in Figure \ref{fig:textcat} (left). As can be seen, the relative prevalence of values is broadly similar to our anntoated sample, though many are predicted to occur with greater frequency, as expected. However, to reiterate, we should be highly skeptical of these findings, given the poor performance of the classifiers, and we can view this analysis to be useful more so to deepen our understanding of appropriate methodology.

\begin{figure}[t!]
\centering
\includegraphics[width=0.4\textwidth]{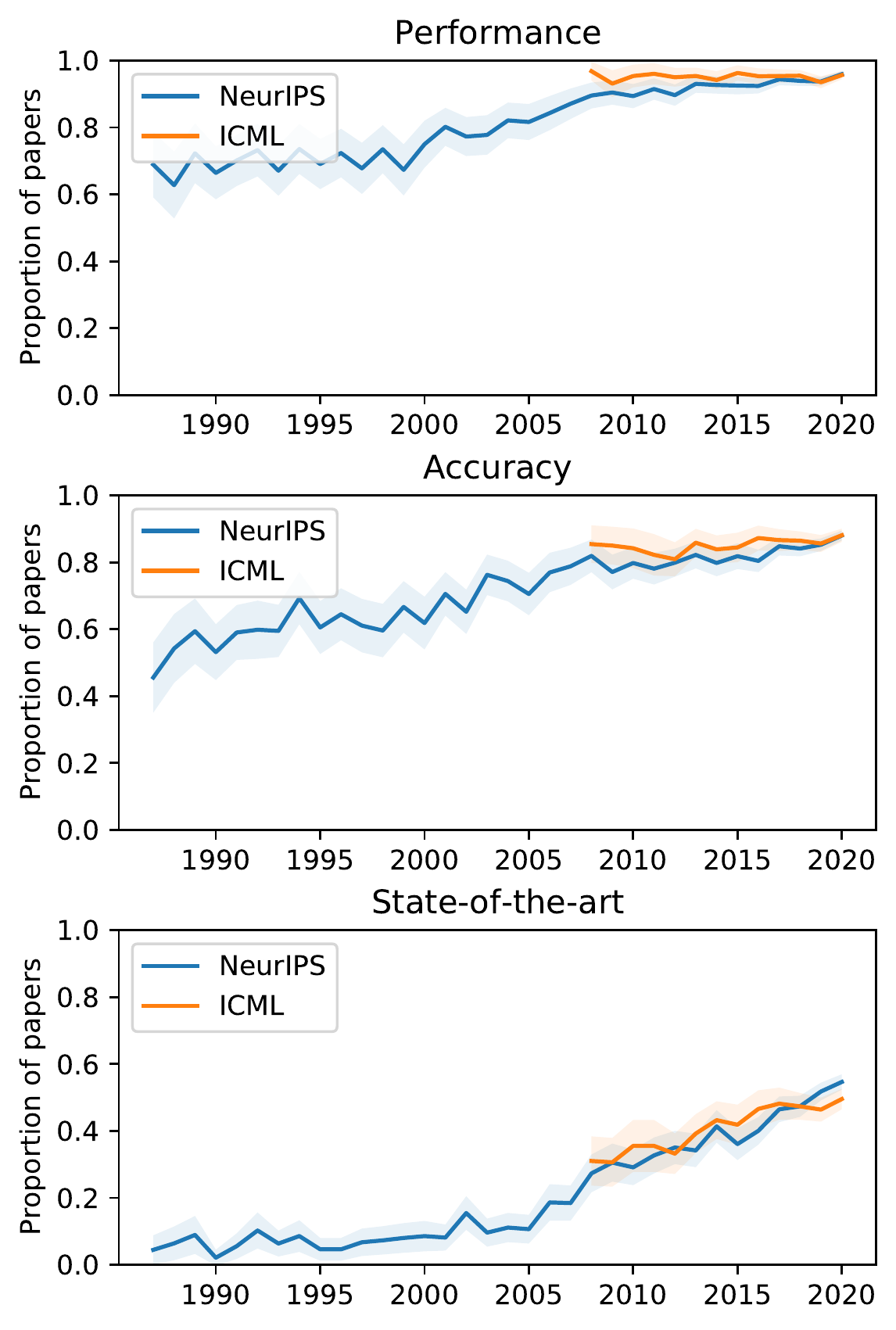}
\caption{Proportion of papers per year (of those published in ICML and NeurIPS) that are classified as having at least one sentence expressing \emph{Performance}, \emph{Accuracy}, or \emph{State-of-the-art}, (top, middle, and bottom), based on simple text classifiers trained on our annotations. Bands show $\pm$2 standard deviations, reflecting the changing overall number of papers per year. }
\label{fig:performance-temporal}
\end{figure}

Finally, as an additional exploration, we focus on the Performance-related values (\emph{Performance}, \emph{Accuracy}, and \emph{State-of-the-art}), which represent the overall most prevalent cluster in our annotations and were relatively easy to identify using classification due to their typically simple and explicit expression. We plot the estimated frequency over time for both conferences. For NeurIPS, which has better archival practices, we extend the analysis back to 1987. We should again treat these results with caution, given all the caveats above, as well as the fact that we are now applying these classifiers outside the temporal range from which the annotations were collected. Nevertheless, the results, shown in Figure \ref{fig:performance-temporal}, suggest that these values have gradually become more common in NeurIPS over time, reinforcing the contingent nature of the dominance of the current set of values. Further investigation is required, however, in order to verify this finding.

\section{Code and Reproducibility}
Our code and annotations are available 
 with a CC BY-NC-SA license at \href{https://github.com/wagnew3/The-Values-Encoded-in-Machine-Learning-Research}{https://github.com/wagnew3/The-Values-Encoded-in-Machine-Learning-Research}. 
The text classification experiments were run on a 2019 Macbook Air.

\section{Potential Negative Societal Impacts}

Because this paper primarily relies on socially conscientious manual annotation of papers already published at NeurIPS and ICML, we believe that the potential negative societal impacts of carrying out these annotations and sharing them are minimal. However, we still briefly comment on this here.

We believe our annotation work poses no risk to living beings, human rights concerns, threats to livelihoods, etc.
Similarly, all annotators are co-authors on this paper, thus there was no risk to participants, beyond what we chose to take on for ourselves. We have further discussed these aspects of the data in \S\ref{sec:data}. Our computational experiments are done locally and have resource usage on par with everyday computer usage.

One area of potential concern to readers, particularly researchers, may be regarding adopting a punitive stance toward individuals, unintentionally casting certain authors in a negative light, or unintentionally contributing to harmful tensions within the ML community. We wish to directly express that throughout this paper we have sought to avoid punitive language toward individuals and adopt language emphasizing systematic patterns. In order to further minimize the former, we have chosen to include randomly selected examples omitting author attributions from quoted sources in the main paper. To complement this and meet the need for completeness, transparency, and reproducibility of our work, we include a full list of cited papers below, so as to acknowledge this work without drawing unnecessary attention to any one particular source.

Although our intention is to broaden and deepen the conversation, we acknowledge that some authors may perceive our work as being not representative of the type of work they would like to see at an ML conference, and possibly detrimental to the conference. However, because of the prominence and influence of machine learning today, it is especially important to have these conversations at these venues, and we hope that our paper will be the basis for useful conversations and future work.
As expressed in the main paper, these perceptions and norms may be precisely those that are more contingent than the community realizes; these norms may be shaped, dismantled, transformed, or reenvisioned for the better. 

\section{Random Examples}
\label{sec:randomexamples}

The list below contains 100 random examples drawn from the annotated data, along with the set of annotated values for each. These sentences were annotated for values within the context of the paper.
\textit{
\begin{itemize}
    \item The problem of minimizing the rank of a matrix variable subject to certain constraints arises in many fields including machine learning, automatic control, and image compression.  {\textbf{Used in practice/Popular}}
	\item Locality-sensitive hashing [6] is an effective technique that performs approximate nearest neighbor searches in time that is sub-linear in the size of the database {\textbf{Approximation}, \textbf{Building on recent work}, \textbf{Effectiveness}, \textbf{Fast}}
	\item In the finite case, analysis of optimization and generalization of fully-trained nets is of course an open problem {\textbf{Formal description/analysis}, \textbf{Generalization}}
	\item So to achieve adversarial robustness, a classifier must generalize in a stronger sense. {\textbf{Generalization}, \textbf{Robustness}}
	\item Robustness to label corruption is similarly improved by wide margins, such that pre-training alone outperforms certain task-specific methods, sometimes even after combining these methods with pre-training. {\textbf{Performance}, \textbf{Robustness}, \textbf{Understanding (for researchers)}}
	\item RBMs have been particularly successful in classification problems either as feature extractors for text and image data (Gehler et al., 2006) or as a good initial
training phase for deep neural network classifiers (Hinton, 2007). {\textbf{Building on recent work}, \textbf{Flexibility/Extensibility}, \textbf{Successful}}
	\item Our theoretical analysis naturally leads to a new formulation
of adversarial defense which has several appealing properties; in particular, it inherits the benefits of scalability to
large datasets exhibited by Tiny ImageNet, and the algorithm achieves state-of-the-art performance on a range of
benchmarks while providing theoretical guarantees. {\textbf{Robustness}, \textbf{Scales up}, \textbf{Security}, \textbf{Theoretical guarantees}}
	\item The current paper focuses on the training loss, but
does not address the test loss. {\textbf{Generalization}}
	\item This result is significant since stochastic methods are highly preferred for their efficiency over deterministic
gradient methods in machine learning applications. {\textbf{Efficiency}}
	\item Ranking, which is to sort objects based on certain factors, is the central problem of applications such as in- formation retrieval (IR) and information ﬁltering. {\textbf{Applies to real world}, \textbf{Used in practice/Popular}}
	\item This subspace is important, because, when projected onto this subspace, the means of the distributions are well-separated, yet the typical distance between points from the same distribution is smaller than in the original space.  {\textbf{Important}}
	\item Overall, the existence of such adversarial examples raises concerns about the robustness of current classifiers. {\textbf{Identifying limitations}, \textbf{Robustness}}
	\item We have shown that biased compressors if naively used can lead to bad generalization, and even non-convergence. {\textbf{Formal description/analysis}, \textbf{Generalization}}
	\item Bartlett and Mendelson [2002] provide a generalization bound for Lipschitz loss functions.  {\textbf{Building on classic work}, \textbf{Generalization}}
	\item The principal advantage of taking this “lateral” approach arises from the fact that compact representation in trajectory space is better motivated physically than compact representation in shape space {\textbf{Realistic world model}}
	\item In this paper, we show that gradient descent on deep overparametrized networks can obtain zero training loss {\textbf{Formal description/analysis}, \textbf{Theoretical guarantees}}
	\item Moreover, web queries often have different meanings for different users (a canonical example is the query jaguar ) suggesting that a ranking with diverse documents may be preferable. {\textbf{Diverse output}, \textbf{User influence}}
	\item We include human performance estimates for all benchmark tasks, which verify that substantial headroom exists between a strong BERT-based baseline and human performance. {\textbf{Learning from humans}, \textbf{Performance}}
	\item Inthis paper we propose a simple and fast algorithm SVP(Singular Value Projec-tion) for rank minimization under affine constraints (ARMP) and show that SVP recovers the minimum rank solution for affine constraints that satisfy a restricted isometry property(RIP). {\textbf{Fast}, \textbf{Novelty}, \textbf{Simplicity}}
	\item We use standard formalization of multiclass classification, where data consists of sample x and its label y
(an integer from 1 to k). {\textbf{Building on classic work}}
	\item A number of recent work has shown that the low rank solution can be recovered exactly via minimizing the trace norm under certain conditions (Recht et al., 2008a; Recht et al., 2008b; Cand\`es and Recht, 2008). {\textbf{Building on recent work}}
	\item This difficulty has necessitated the use of a heuristic inference procedure, that
nonetheless was accurate enough for successful learning.  {\textbf{Accuracy}, \textbf{Successful}}
	\item We illustrate such potential by measuring search space properties relevant to architecture search. {\textbf{Quantitative evidence (e.g. experiments)}}
	\item Deep architectures consist of feature detector units arranged in layers. Lower layers detect simple features and feed into higher layers, which in turn detect more complex features.  {\textbf{Simplicity}}
	\item This makes the updates hard to massively parallelize at a coarse, dataparallel level (e.g., by computing the updates in parallel and summing them together centrally) without losing the critical stochastic nature of the updates.  {\textbf{Large scale}, \textbf{Parallelizability / distributed}}
	\item This suggests future work on model robustness should evaluate proposed methods with pretraining in order to correctly gauge their utility, and some work could specialize pre-training for these downstream tasks. {\textbf{Robustness}}
	\item Adversarial training remains among the most trusted defenses, but it is nearly intractable on largescale problems.  {\textbf{Scales up}, \textbf{Security}}
	\item For complex robots such as humanoids or light-weight arms, it is often hard to model the system sufficiently well and, thus, modern regression methods offer a viable alternative [7,8]. {\textbf{Realistic world model}}
	\item In
contrast to prior work that operates in this goal-setting model, we use states as goals directly, which
allows for simple and fast training of the lower layer. {\textbf{Reduced training time}, \textbf{Simplicity}}
	\item Meanwhile, using less resources tends to produce less compelling results (Negrinho and Gordon, 2017; Baker et al., 2017a).  {\textbf{Requires few resources}}
	\item This finding represents an exciting opportunity for defense against neural fake news: the best models for generating neural disinformation are also the best models at detecting it. {\textbf{Applies to real world}}
	\item Our strong empirical results suggest that
randomized smoothing is a promising direction
for future research into adversarially robust classification.  {\textbf{Quantitative evidence (e.g. experiments)}, \textbf{Robustness}, \textbf{Security}}
	\item We then turn our attention to identifying the roots of BatchNorm’s success. {\textbf{Successful}, \textbf{Understanding (for researchers)}}
	\item We also report the results of large-scale experiments comparing these three methods which demonstrate the benefits of the mixture weight method: this method consumes less resources, while achieving a performance comparable to that of standard approaches. {\textbf{Large scale}, \textbf{Performance}, \textbf{Requires few resources}}
	\item This paper does not cover the the generalization of over-parameterized neural networks to the test data. {\textbf{Avoiding train/test discrepancy}, \textbf{Generalization}}
	\item While there has been success with robust classifiers on simple datasets [31, 36, 44, 48], more complicated datasets still exhibit a large gap between “‘standard” and robust accuracy [3, 11]. {\textbf{Applies to real world}, \textbf{Robustness}, \textbf{Successful}}
	\item In this paper, we have shown theoretically how independence between examples can make the actual effect much smaller.  {\textbf{Novelty}, \textbf{Theoretical guarantees}}
	\item We provide empirical evidence that several recently-used methods for estimating the probability of held-out documents are inaccurate and can change the results of model comparison. {\textbf{Accuracy}, \textbf{Building on recent work}, \textbf{Quantitative evidence (e.g. experiments)}}
	\item This agreement is robust across different architectures, optimization methods, and loss functions {\textbf{Robustness}}
	\item Unfortunately, due to the slow-changing policy in an actor-critic setting, the current and target value estimates remain too similar to avoid maximization bias. {\textbf{Accuracy}}
	\item As a future work, we are pursuing a better understanding of probabilistic distributions on the Grassmann manifold. {\textbf{Understanding (for researchers)}}
	\item We also view these results as an opportunity to encourage the community to pursue a more systematic investigation of the algorithmic toolkit of deep learning and the underpinnings of its effectiveness. {\textbf{Effectiveness}, \textbf{Understanding (for researchers)}}
	\item This challenge is further exacerbated in continuous state and action spaces, where a separate actor network is often used to perform the maximization in Q-learning.  {\textbf{Performance}}
	\item The vulnerability of neural networks to adversarial perturbations has recently been a source of much discussion and is still poorly understood. {\textbf{Robustness}, \textbf{Understanding (for researchers)}}
	\item Most of the evaluation methods described in this paper extend readily to more complicated topic models— including non-parametric versions based on hierarchical Dirichlet processes (Teh et al., 2006)—since they only require a MCMC algorithm for sampling the latent topic assignments z for each document and a way of evaluating probability P(w | z, $\Phi$, $\alpha$m). {\textbf{Flexibility/Extensibility}, \textbf{Understanding (for researchers)}}
	\item In a formulation closely related to the dual problem, we have: $\hat{w}$ = argmin w:F (w)$\leq$c 1 n Xn i=1 $\ell$(hw, xii, yi) (2) where, instead of regularizing, a hard restriction over the parameter space is imposed (by the constant c).  {\textbf{Formal description/analysis}}
	\item Second, we evaluate a surrogate loss function from four aspects: (a) consistency, (b) soundness, (c) mathemat- ical properties of continuity, diﬀerentiability, and con- vexity, and (d) computational eﬃciency in learning. {\textbf{Efficiency}}
	\item This leads to two natural questions that we try to answer in this paper: (1) Is it feasible to perform optimization in this very large feature space with cost which is polynomial in the size of the input space?  {\textbf{Performance}}
	\item Despite its pervasiveness, the exact reasons for BatchNorm’s effectiveness are still poorly understood. {\textbf{Understanding (for researchers)}}
	\item We have presented confidenceweighted linear classifiers, a new learning method designed for NLP problems based on the notion of parameter confidence.  {\textbf{Novelty}}
	\item In addition, the experiments reported here suggest that (like other strategies recently proposed to train deep deterministic or stochastic neural networks) the curriculum strategies appear on the surface to operate like a regularizer, i.e., their beneficial effect is most pronounced on the test set.  {\textbf{Beneficence}, \textbf{Quantitative evidence (e.g. experiments)}}
	\item These
give further inside into hash-spaces and explain previously
made empirical observations. {\textbf{Understanding (for researchers)}}
	\item This means that current algorithms reach their limit at problems of size 1TB whenever the algorithm is I/O bound (this amounts to a training time of 3 hours), or even smaller problems whenever the model parametrization makes the algorithm CPU bound. {\textbf{Memory efficiency}, \textbf{Reduced training time}}
	\item Much of the results presented were based on the assumption that the target distribution is some mixture of the source distributions.  {\textbf{Valid assumptions}}
	\item Empirical investigation revealed that this agrees well with actual training dynamics and predictive distributions across fully-connected, convolutional, and even wide residual network architectures, as well as with different optimizers (gradient descent, momentum, mini-batching) and loss functions (MSE, cross-entropy). {\textbf{Generalization}, \textbf{Quantitative evidence (e.g. experiments)}, \textbf{Understanding (for researchers)}}
	\item We design a new spectral norm that encodes this a priori assumption, without the prior knowledge of the partition of tasks into groups, resulting
in a new convex optimization formulation for multi-task learning.  {\textbf{Novelty}}
	\item Recent progress in natural language generation has raised dual-use concerns.  {\textbf{Progress}}
	\item These kernel functions can be used
in shallow architectures, such as support vector machines (SVMs), or in deep
kernel-based architectures that we call multilayer kernel machines (MKMs). {\textbf{Flexibility/Extensibility}}
	\item Using MCMC instead of variational methods for approximate inference in Bayesian matrix factorization models leads to much larger improvements over the MAP trained models, which suggests that the assumptions made by the variational methods about the structure of the posterior are not entirely reasonable. {\textbf{Understanding (for researchers)}}
	\item In particular, the deep belief network (DBN) (Hinton et al., 2006) is a multilayer generative model where each layer encodes statistical dependencies among the units in the layer below it; it is trained to (approximately) maximize the likelihood of its training data. {\textbf{Approximation}, \textbf{Data efficiency}}
	\item Furthermore,  the learning accuracy and performance of our LGP approach will be compared with other important standard methods in Section 4, e.g., LWPR [8], standard GPR [1], sparse online Gaussian process regression (OGP) [5] and $\nu$-support vector regression ($\nu$-SVR) [11], respectively {\textbf{Accuracy}, \textbf{Performance}, \textbf{Quantitative evidence (e.g. experiments)}}
	\item • propose a simple method based on weighted minibatches to stochastically train with arbitrary weights on the terms of our decomposition without any additional hyperparameters. {\textbf{Efficiency}, \textbf{Simplicity}}
	\item For example, Ng (2004) examined the task of PAC learning a sparse predictor and analyzed cases in which an $\ell$1 constraint results in better solutions than an $\ell$2 constraint. {\textbf{Building on recent work}}
	\item Graph Convolutional Networks (GCNs) (Kipf and Welling, 2017) are an efficient variant of Convolutional Neural Networks (CNNs) on graphs. GCNs stack layers of learned first-order spectral filters followed by a nonlinear activation function to learn graph representations.  {\textbf{Efficiency}}
	\item This is a linear convergence rate. {\textbf{Building on recent work}, \textbf{Efficiency}, \textbf{Quantitative evidence (e.g. experiments)}, \textbf{Theoretical guarantees}}
	\item However, as we
observe more interactions, this could emerge as a clear feature. {\textbf{Building on recent work}, \textbf{Data efficiency}}
	\item Here we propose the first method that supports arbitrary low accuracy and even biased
compression operators, such as in (Alistarh et al., 2018; Lin et al., 2018; Stich et al., 2018). {\textbf{Accuracy}, \textbf{Novelty}}
	\item Much recent work has been done on understanding under what conditions we can learn a mixture model. {\textbf{Understanding (for researchers)}}
	\item For this reason, we present an extension of the standard greedy OMP algorithm that can be applied to general struc- tured sparsity problems, and more importantly, meaningful sparse recovery bounds can be obtained for this algorithm. {\textbf{Building on recent work}}
	\item In this paper we show that this assumption is indeed necessary: by considering a simple yet prototypical exampleof GAN training we analytically show that (unregularized) GAN training is not always locally convergent {\textbf{Formal description/analysis}}
	\item Overestimation bias is a property of Q-learning in which the maximization of a noisy value estimate induces a consistent overestimation {\textbf{Accuracy}}
	\item This drawback prevents GPR from applications which need large amounts of training data and require fast computation, e.g., online learning of inverse dynamics model for model-based robot control {\textbf{Fast}, \textbf{Large scale}}
	\item This is problematic since we find there are techniques which do not comport well with pre-training; thus some evaluations of robustness are less representative of real-world performance than previously thought. {\textbf{Applies to real world}, \textbf{Performance}, \textbf{Robustness}}
	\item Approximation of this prior structure through simple, efficient hyperparameter optimization steps is sufficient to achieve these performance gains {\textbf{Approximation}, \textbf{Efficiency}, \textbf{Performance}, \textbf{Simplicity}}
	\item The second mysterious phenomenon in training deep neural networks is “deeper networks are harder to train.” {\textbf{Performance}}
	\item However, the definition of our metric is sufficiently general that it could easily be used to test, for example, invariance of auditory features to rate of speech, or invariance of textual features to author identity. {\textbf{Generalization}}
	\item In Sec. 6 we test the proposed algorithm for
face recognition and object categorization tasks. {\textbf{Applies to real world}, \textbf{Quantitative evidence (e.g. experiments)}}
	\item It is possible to train classification
RBMs directly for classification performance; the gradient is fairly simple and certainly tractable. {\textbf{Performance}}
	\item Figure 1 contrasts these two approaches. Defining and evaluating models using ODE solvers has several benefits: {\textbf{Beneficence}}
	\item They claim to achieve 12\% robustness against non-targeted attacks that are within an
`2 radius of 3 (for images with pixels in [0, 1]). {\textbf{Generalization}, \textbf{Robustness}}
	\item Two commonly used penalties are the 1- norm and the square of the 2-norm of w. {\textbf{Used in practice/Popular}}
	\item What should platforms do? Video-sharing platforms like YouTube use deep neural networks to scan videos while they are uploaded, to filter out content like pornography (Hosseini et al., 2017). {\textbf{Applies to real world}}
	\item We mention various properties
of this penalty, and provide conditions for the consistency
of support estimation in the regression setting. Finally, we
report promising results on both simulated and real data {\textbf{Applies to real world}}
	\item There
could be a separate feature for “high school student,” “male,” “athlete,” and “musician” and the
presence or absence of each of these features is what defines each person and determines their
relationships. {\textbf{Building on recent work}}
	\item So, the over-parameterized convergence theory of DNN is much simpler than that of RNN. {\textbf{Simplicity}, \textbf{Understanding (for researchers)}}
	\item Other threat models are possible: for instance, an adversary might generate comments or have entire dialogue agents, they might start with a human-written news article and modify a few sentences, and they might fabricate images or video.  {\textbf{Learning from humans}}
	\item More generally, we hope that future work will be able to avoid relying on obfuscated gradients (and other methods that only prevent gradient descent-based attacks) for perceived robustness, and use our evaluation approach to detect when this occurs.  {\textbf{Generality}, \textbf{Robustness}}
	\item For example, the learned linear combination does not consistently outperform either the uniform combination of base kernels or simply the best single base kernel (see, for example, UCI dataset experiments in [9, 12], see also NIPS 2008 workshop). {\textbf{Performance}}
	\item Our main contributions are: • We analyze GP-UCB, an intuitive algorithm for GP optimization, when the function is either sampled from a known GP, or has low RKHS norm. {\textbf{Optimal}}
	\item For the standard linear setting, Dani et al. (2008) provide a near-complete characterization explicitly dependent on the dimensionality. In the GP setting, the challenge is to characterize complexity in a different manner, through properties of the kernel function.  {\textbf{Building on classic work}}
	\item This allows us to map each architecture A to its approximate hyperparameter optimized accuracy {\textbf{Accuracy}}
	\item Unfortunately, they could only apply their method to linear networks.  {\textbf{Flexibility/Extensibility}}
	\item The strength of the adversary then allows for a trade-off between the enforced prior, and the data-dependent features. {\textbf{Understanding (for researchers)}}
	\item We observe that the computational bottleneck of NAS is the training of each child model to convergence, only to measure its accuracy whilst throwing away all the trained weights.  {\textbf{Accuracy}}
	\item We show that the number of subproblems need only be logarithmic in the total number of possible labels, making thisapproach radically more efficient than others. {\textbf{Efficiency}}
	\item We establish a new
notion of quadratic approximation of the neural network, and connect it to the
SGD theory of escaping saddle points. {\textbf{Novelty}, \textbf{Unifying ideas or integrating components}}
	\item In this work, we decompose the prediction error for adversarial
examples (robust error) as the sum of the natural
(classification) error and boundary error, and provide a differentiable upper bound using the theory
of classification-calibrated loss, which is shown to
be the tightest possible upper bound uniform over
all probability distributions and measurable predictors. {\textbf{Accuracy}, \textbf{Robustness}, \textbf{Theoretical guarantees}}
	\item  A limit on the number of queries can be a result
of limits on other resources, such as a time limit if inference time is a bottleneck or a monetary limit if the
attacker incurs a cost for each query. {\textbf{Applies to real world}, \textbf{Low cost}, \textbf{Requires few resources}}
	\item Preliminary experiments demonstrate that it is significantly faster than batch alternatives on large datasets that may contain millions of training examples, yet it does not require learning rate tuning like regular stochastic gradient descent methods.  {\textbf{Quantitative evidence (e.g. experiments)}, \textbf{Reduced training time}}
	\item SuperGLUE is available at super.gluebenchmark.com. {\textbf{Facilitating use (e.g. sharing code)}}
\end{itemize}}

\section{Full List of Cited Papers}
\label{sec:paperlist}

The full list of annotated papers is given below, along with the annotated scores (in square brackets) for \emph{Discussion of Negative Potential} [left] (0 = Doesn’t mention negative potential; 1 = Mentions but does not discuss negative potential; 2 = Discusses negative potential) and \emph{Justification} [right] (1 = Does not mention societal need; 2 = States but does not justify how it connects to a societal need; 3 = States and somewhat justifies how it connects to a societal need; 4 = States and rigorously justifies how it connects to a a societal need). Note that due to minor errors in the data sources used, the distribution of papers over venues and years is not perfectly balanced. For the same reason, the list also contains one paper from 2010 (rather than 2009), as well as one paper that was retracted before publication at NeurIPS (marked with a $^*$).

\begin{itemize}
	\item Mingxing Tan, Quoc Le. \href{http://proceedings.mlr.press/v97/tan19a.html}{EfficientNet: Rethinking Model Scaling for Convolutional Neural Networks.} In \emph{Proceedings of ICML}, 2019. [0/1]
	\item Sanjeev Arora, Simon Du, Wei Hu, Zhiyuan Li, Ruosong Wang. \href{http://proceedings.mlr.press/v97/arora19a.html}{Fine-Grained Analysis of Optimization and Generalization for Overparameterized Two-Layer Neural Networks.} In \emph{Proceedings of ICML}, 2019. [0/1]
	\item Jeremy Cohen, Elan Rosenfeld, Zico Kolter. \href{http://proceedings.mlr.press/v97/cohen19c.html}{Certified Adversarial Robustness via Randomized Smoothing.} In \emph{Proceedings of ICML}, 2019. [0/1]
	\item Hongyang Zhang, Yaodong Yu, Jiantao Jiao, Eric Xing, Laurent El Ghaoui, Michael Jordan. \href{http://proceedings.mlr.press/v97/zhang19p.html}{Theoretically Principled Trade-off between Robustness and Accuracy.} In \emph{Proceedings of ICML}, 2019. [0/2]
	\item Kaitao Song, Xu Tan, Tao Qin, Jianfeng Lu, Tie-Yan Liu. \href{http://proceedings.mlr.press/v97/song19d.html}{MASS: Masked Sequence to Sequence Pre-training for Language Generation.} In \emph{Proceedings of ICML}, 2019. [0/1]
	\item Felix Wu, Amauri Souza, Tianyi Zhang, Christopher Fifty, Tao Yu, Kilian Weinberger. \href{http://proceedings.mlr.press/v97/wu19e.html}{Simplifying Graph Convolutional Networks.} In \emph{Proceedings of ICML}, 2019. [0/1]
	\item Benjamin Recht, Rebecca Roelofs, Ludwig Schmidt, Vaishaal Shankar. \href{http://proceedings.mlr.press/v97/recht19a.html}{Do ImageNet Classifiers Generalize to ImageNet?} In \emph{Proceedings of ICML}, 2019. [0/2]
	\item Justin Gilmer, Nicolas Ford, Nicholas Carlini, Ekin Cubuk. \href{http://proceedings.mlr.press/v97/gilmer19a.html}{Adversarial Examples Are a Natural Consequence of Test Error in Noise.} In \emph{Proceedings of ICML}, 2019. [0/1]
	\item Chris Ying, Aaron Klein, Eric Christiansen, Esteban Real, Kevin Murphy, Frank Hutter. \href{http://proceedings.mlr.press/v97/ying19a.html}{NAS-Bench-101: Towards Reproducible Neural Architecture Search.} In \emph{Proceedings of ICML}, 2019. [0/2]
	\item Dan Hendrycks, Kimin Lee, Mantas Mazeika. \href{http://proceedings.mlr.press/v97/hendrycks19a.html}{Using Pre-Training Can Improve Model Robustness and Uncertainty.} In \emph{Proceedings of ICML}, 2019. [0/1]
	\item Sai Praneeth Karimireddy, Quentin Rebjock, Sebastian Stich, Martin Jaggi. \href{http://proceedings.mlr.press/v97/karimireddy19a.html}{Error Feedback Fixes SignSGD and other Gradient Compression Schemes.} In \emph{Proceedings of ICML}, 2019. [0/1]
	\item Anastasia Koloskova, Sebastian Stich, Martin Jaggi. \href{http://proceedings.mlr.press/v97/koloskova19a.html}{Decentralized Stochastic Optimization and Gossip Algorithms with Compressed Communication.} In \emph{Proceedings of ICML}, 2019. [0/2]
	\item Han Zhang, Ian Goodfellow, Dimitris Metaxas, Augustus Odena. \href{http://proceedings.mlr.press/v97/zhang19d.html}{Self-Attention Generative Adversarial Networks.} In \emph{Proceedings of ICML}, 2019. [0/1]
	\item Zeyuan Allen-Zhu, Yuanzhi Li, Zhao Song. \href{http://proceedings.mlr.press/v97/allen-zhu19a.html}{A Convergence Theory for Deep Learning via Over-Parameterization.} In \emph{Proceedings of ICML}, 2019. [0/1]
	\item Simon Du, Jason Lee, Haochuan Li, Liwei Wang, Xiyu Zhai. \href{http://proceedings.mlr.press/v97/du19c.html}{Gradient Descent Finds Global Minima of Deep Neural Networks.} In \emph{Proceedings of ICML}, 2019. [0/1]
	\item Anish Athalye, Nicholas Carlini, David Wagner. \href{http://proceedings.mlr.press/v80/athalye18a.html}{Obfuscated Gradients Give a False Sense of Security: Circumventing Defenses to Adversarial Examples.} In \emph{Proceedings of ICML}, 2018. [0/2]
	\item Hieu Pham, Melody Guan, Barret Zoph, Quoc Le, Jeff Dean. \href{http://proceedings.mlr.press/v80/pham18a.html}{Efficient Neural Architecture Search via Parameters Sharing.} In \emph{Proceedings of ICML}, 2018. [0/1]
	\item Tuomas Haarnoja, Aurick Zhou, Pieter Abbeel, Sergey Levine. \href{http://proceedings.mlr.press/v80/haarnoja18b.html}{Soft Actor-Critic: Off-Policy Maximum Entropy Deep Reinforcement Learning with a Stochastic Actor.} In \emph{Proceedings of ICML}, 2018. [0/2]
	\item Lasse Espeholt, Hubert Soyer, Remi Munos, Karen Simonyan, Vlad Mnih, Tom Ward, Yotam Doron, Vlad Firoiu, Tim Harley, Iain Dunning, Shane Legg, Koray Kavukcuoglu. \href{http://proceedings.mlr.press/v80/espeholt18a.html}{IMPALA: Scalable Distributed Deep-RL with Importance Weighted Actor-Learner Architectures.} In \emph{Proceedings of ICML}, 2018. [0/1]
	\item Scott Fujimoto, Herke Hoof, David Meger. \href{http://proceedings.mlr.press/v80/fujimoto18a.html}{Addressing Function Approximation Error in Actor-Critic Methods.} In \emph{Proceedings of ICML}, 2018. [0/1]
	\item Hyunjik Kim, Andriy Mnih. \href{http://proceedings.mlr.press/v80/kim18b.html}{Disentangling by Factorising.} In \emph{Proceedings of ICML}, 2018. [0/0]
	\item Lars Mescheder, Andreas Geiger, Sebastian Nowozin. \href{http://proceedings.mlr.press/v80/mescheder18a.html}{Which Training Methods for GANs do actually Converge?} In \emph{Proceedings of ICML}, 2018. [0/1]
	\item Sanjeev Arora, Rong Ge, Behnam Neyshabur, Yi Zhang. \href{http://proceedings.mlr.press/v80/arora18b.html}{Stronger generalization bounds for deep nets via a compression approach.} In \emph{Proceedings of ICML}, 2018. [0/3]
	\item Andrew Ilyas, Logan Engstrom, Anish Athalye, Jessy Lin. \href{http://proceedings.mlr.press/v80/ilyas18a.html}{Black-box Adversarial Attacks with Limited Queries and Information.} In \emph{Proceedings of ICML}, 2018. [0/2]
	\item Niranjan Srinivas, Andreas Krause, Sham Kakade, Matthias Seeger. \href{https://icml.cc/Conferences/2010/papers/422.pdf}{Gaussian Process Optimization in the Bandit Setting: No Regret and Experimental Design.} In \emph{Proceedings of ICML}, 2010. [0/1]
	\item Honglak Lee, Roger Grosse, Rajesh Ranganath and Andrew Ng. \href{https://icml.cc/Conferences/2009/papers/571.pdf}{Convolutional deep belief networks for scalable unsupervised learning of hierarchical representations.} In \emph{Proceedings of ICML}, 2009. [0/1]
	\item Julien Mairal, Francis Bach, Jean Ponce and Guillermo Sapiro. \href{https://icml.cc/Conferences/2009/papers/364.pdf}{Online dictionary learning for sparse coding.} In \emph{Proceedings of ICML}, 2009. [0/1]
	\item Yoshua Bengio, Jerome Louradour, Ronan Collobert and Jason Weston. \href{https://icml.cc/Conferences/2009/papers/119.pdf}{Curriculum learning.} In \emph{Proceedings of ICML}, 2009. [0/1]
	\item Laurent Jacob, Guillaume Obozinski and Jean-Philippe Vert. \href{https://icml.cc/Conferences/2009/papers/471.pdf}{Group Lasso with Overlaps and Graph Lasso.} In \emph{Proceedings of ICML}, 2009. [0/3]
	\item Chun-Nam Yu and Thorsten Joachims. \href{https://icml.cc/Conferences/2009/papers/420.pdf}{Learning structural SVMs with latent variables.} In \emph{Proceedings of ICML}, 2009. [0/2]
	\item Kilian Weinberger, Anirban Dasgupta, Josh Attenberg, John Langford and Alex Smola. \href{https://icml.cc/Conferences/2009/papers/407.pdf}{Feature hashing for large scale multitask learning.} In \emph{Proceedings of ICML}, 2009. [0/2]
	\item Hanna Wallach, Iain Murray, Ruslan Salakhutdinov, and David Mimno. \href{https://icml.cc/Conferences/2009/papers/356.pdf}{Evaluation methods for topic models.} In \emph{Proceedings of ICML}, 2009. [0/1]
	\item Kamalika Chaudhuri, Sham Kakade, Karen Livescu and Karthik Sridharan. \href{https://icml.cc/Conferences/2009/papers/317.pdf}{Multi-view clustering via canonical correlation analysis.} In \emph{Proceedings of ICML}, 2009. [0/2]
	\item Shuiwang Ji and Jieping Ye. \href{https://icml.cc/Conferences/2009/papers/151.pdf}{An accelerated gradient method for trace norm minimization.} In \emph{Proceedings of ICML}, 2009. [0/3]
	\item Junzhou Huang, Tong Zhang and Dimitris Metaxas. \href{https://icml.cc/Conferences/2009/papers/452.pdf}{Learning with structured sparsity.} In \emph{Proceedings of ICML}, 2009. [0/1]
	\item Rajat Raina, Anand Madhavan and Andrew Ng. \href{https://icml.cc/Conferences/2009/papers/218.pdf}{Large-scale deep unsupervised learning using graphics processors.} In \emph{Proceedings of ICML}, 2009. [0/2]
	\item Ronan Collobert and Jason Weston. \href{https://icml.cc/Conferences/2008/papers/391.pdf}{A unified architecture for natural language processing: deep neural networks with multitask learning.} In \emph{Proceedings of ICML}, 2008. [0/2]
	\item Pascal Vincent, Hugo Larochelle, Yoshua Bengio, and Pierre-Antoine Manzagol. \href{https://icml.cc/Conferences/2008/papers/592.pdf}{Extracting and composing robust features with denoising autoencoders.} In \emph{Proceedings of ICML}, 2008. [0/1]
	\item Ruslan Salakhutdinov and Andriy Mnih. \href{https://icml.cc/Conferences/2008/papers/600.pdf}{Bayesian probabilistic matrix factorization using Markov chain Monte Carlo.} In \emph{Proceedings of ICML}, 2008. [0/1]
	\item John Duchi, Shai Shalev-Shwartz, Yoram Singer, and Tushar Chandra. \href{https://icml.cc/Conferences/2008/papers/361.pdf}{Efficient projections onto the l1-ball for learning in high dimensions.} In \emph{Proceedings of ICML}, 2008. [0/1]
	\item Cho-Jui Hsieh, Kai-Wei Chang, Chih-Jen Lin, S. Sathiya Keerthi, and S. Sundararajan. \href{https://icml.cc/Conferences/2008/papers/166.pdf}{A dual coordinate descent method for large-scale linear SVM.} In \emph{Proceedings of ICML}, 2008. [0/1]
	\item Tijmen Tieleman. \href{https://icml.cc/Conferences/2008/papers/638.pdf}{Training restricted Boltzmann machines using approximations to the likelihood gradient.} In \emph{Proceedings of ICML}, 2008. [0/1]
	\item Hugo Larochelle and Yoshua Bengio. \href{https://icml.cc/Conferences/2008/papers/601.pdf}{Classification using discriminative restricted Boltzmann machines.} In \emph{Proceedings of ICML}, 2008. [0/1]
	\item Jihun Hamm and Daniel Lee. \href{https://icml.cc/Conferences/2008/papers/312.pdf}{Grassmann discriminant analysis: a unifying view on subspace-based learning.} In \emph{Proceedings of ICML}, 2008. [0/1]
	\item Fen Xia, Tie-Yan Liu, Jue Wang, Wensheng Zhang, and Hang Li. \href{https://icml.cc/Conferences/2008/papers/167.pdf}{Listwise Approach to Learning to Rank - Theory and Algorithm.} In \emph{Proceedings of ICML}, 2008. [0/1]
	\item Filip Radlinski, Robert Kleinberg, and Thorsten Joachims. \href{https://icml.cc/Conferences/2008/papers/264.pdf}{Learning diverse rankings with multi-armed bandits.} In \emph{Proceedings of ICML}, 2008. [0/1]
	\item Mark Dredze, Koby Crammer, and Fernando Pereira. \href{https://icml.cc/Conferences/2008/papers/322.pdf}{Confidence-weighted linear classification.} In \emph{Proceedings of ICML}, 2008. [0/1]
	\item Ruslan Salakhutdinov and Iain Murray. \href{https://icml.cc/Conferences/2008/papers/573.pdf}{On the quantitative analysis of deep belief networks.} In \emph{Proceedings of ICML}, 2008. [0/1]
	\item Zhilin Yang, Zihang Dai, Yiming Yang, Jaime Carbonell, Russ R. Salakhutdinov, Quoc V. Le. \href{https://proceedings.neurips.cc/paper/2019/hash/dc6a7e655d7e5840e66733e9ee67cc69-Abstract.html}{XLNet: Generalized Autoregressive Pretraining for Language Understanding.} In \emph{Proceedings of NeurIPS}, 2019. [0/1]
	\item Alexis CONNEAU, Guillaume Lample. \href{https://proceedings.neurips.cc/paper/2019/hash/c04c19c2c2474dbf5f7ac4372c5b9af1-Abstract.html}{Cross-lingual Language Model Pretraining.} In \emph{Proceedings of NeurIPS}, 2019. [0/4]
	\item Andrew Ilyas, Shibani Santurkar, Dimitris Tsipras, Logan Engstrom, Brandon Tran, Aleksander Madry. \href{https://proceedings.neurips.cc/paper/2019/hash/e2c420d928d4bf8ce0ff2ec19b371514-Abstract.html}{Adversarial Examples Are Not Bugs, They Are Features.} In \emph{Proceedings of NeurIPS}, 2019. [0/1]
	\item Jaehoon Lee, Lechao Xiao, Samuel Schoenholz, Yasaman Bahri, Roman Novak, Jascha Sohl-Dickstein, Jeffrey Pennington. \href{https://proceedings.neurips.cc/paper/2019/hash/0d1a9651497a38d8b1c3871c84528bd4-Abstract.html}{Wide Neural Networks of Any Depth Evolve as Linear Models Under Gradient Descent.} In \emph{Proceedings of NeurIPS}, 2019. [0/1]
	\item David Berthelot, Nicholas Carlini, Ian Goodfellow, Nicolas Papernot, Avital Oliver, Colin A. Raffel. \href{https://proceedings.neurips.cc/paper/2019/hash/1cd138d0499a68f4bb72bee04bbec2d7-Abstract.html}{MixMatch: A Holistic Approach to Semi-Supervised Learning.} In \emph{Proceedings of NeurIPS}, 2019. [0/1]
	\item Adam Paszke, Sam Gross, Francisco Massa, Adam Lerer, James Bradbury, Gregory Chanan, Trevor Killeen, Zeming Lin, Natalia Gimelshein, Luca Antiga, Alban Desmaison, Andreas Kopf, Edward Yang, Zachary DeVito, Martin Raison, Alykhan Tejani, Sasank Chilamkurthy, Benoit Steiner, Lu Fang, Junjie Bai, Soumith Chintala. \href{https://proceedings.neurips.cc/paper/2019/hash/bdbca288fee7f92f2bfa9f7012727740-Abstract.html}{PyTorch: An Imperative Style, High-Performance Deep Learning Library.} In \emph{Proceedings of NeurIPS}, 2019. [0/1]
	\item Sanjeev Arora, Simon S. Du, Wei Hu, Zhiyuan Li, Russ R. Salakhutdinov, Ruosong Wang. \href{https://proceedings.neurips.cc/paper/2019/hash/dbc4d84bfcfe2284ba11beffb853a8c4-Abstract.html}{On Exact Computation with an Infinitely Wide Neural Net.} In \emph{Proceedings of NeurIPS}, 2019. [0/1]
	\item Li Dong, Nan Yang, Wenhui Wang, Furu Wei, Xiaodong Liu, Yu Wang, Jianfeng Gao, Ming Zhou, Hsiao-Wuen Hon. \href{https://proceedings.neurips.cc/paper/2019/hash/c20bb2d9a50d5ac1f713f8b34d9aac5a-Abstract.html}{Unified Language Model Pre-training for Natural Language Understanding and Generation.} In \emph{Proceedings of NeurIPS}, 2019. [0/1]
	\item Ali Shafahi, Mahyar Najibi, Mohammad Amin Ghiasi, Zheng Xu, John Dickerson, Christoph Studer, Larry S. Davis, Gavin Taylor, Tom Goldstein. \href{https://proceedings.neurips.cc/paper/2019/hash/7503cfacd12053d309b6bed5c89de212-Abstract.html}{Adversarial Training for Free!} In \emph{Proceedings of NeurIPS}, 2019. [0/3]
	\item Jiasen Lu, Dhruv Batra, Devi Parikh, Stefan Lee. \href{https://proceedings.neurips.cc/paper/2019/hash/c74d97b01eae257e44aa9d5bade97baf-Abstract.html}{ViLBERT: Pretraining Task-Agnostic Visiolinguistic Representations for Vision-and-Language Tasks.} In \emph{Proceedings of NeurIPS}, 2019. [0/1]
	\item Alex Wang, Yada Pruksachatkun, Nikita Nangia, Amanpreet Singh, Julian Michael, Felix Hill, Omer Levy, Samuel Bowman. \href{https://proceedings.neurips.cc/paper/2019/hash/4496bf24afe7fab6f046bf4923da8de6-Abstract.html}{SuperGLUE: A Stickier Benchmark for General-Purpose Language Understanding Systems.} In \emph{Proceedings of NeurIPS}, 2019. [1/1]
	\item Rowan Zellers, Ari Holtzman, Hannah Rashkin, Yonatan Bisk, Ali Farhadi, Franziska Roesner, Yejin Choi. \href{https://proceedings.neurips.cc/paper/2019/hash/3e9f0fc9b2f89e043bc6233994dfcf76-Abstract.html}{Defending Against Neural Fake News.} In \emph{Proceedings of NeurIPS}, 2019. [2/4]
	\item Yuan Cao, Quanquan Gu. \href{https://proceedings.neurips.cc/paper/2019/hash/cf9dc5e4e194fc21f397b4cac9cc3ae9-Abstract.html}{Generalization Bounds of Stochastic Gradient Descent for Wide and Deep Neural Networks.} In \emph{Proceedings of NeurIPS}, 2019. [0/1]
	\item Florian Tramer, Dan Boneh. \href{https://proceedings.neurips.cc/paper/2019/hash/5d4ae76f053f8f2516ad12961ef7fe97-Abstract.html}{Adversarial Training and Robustness for Multiple Perturbations.} In \emph{Proceedings of NeurIPS}, 2019. [0/2]
	\item Yair Carmon, Aditi Raghunathan, Ludwig Schmidt, John C. Duchi, Percy S. Liang. \href{https://proceedings.neurips.cc/paper/2019/hash/32e0bd1497aa43e02a42f47d9d6515ad-Abstract.html}{Unlabeled Data Improves Adversarial Robustness.} In \emph{Proceedings of NeurIPS}, 2019. [0/1]
	\item Lars Maaløe, Marco Fraccaro, Valentin Liévin, Ole Winther. \href{https://proceedings.neurips.cc/paper/2019/hash/9bdb8b1faffa4b3d41779bb495d79fb9-Abstract.html}{BIVA: A Very Deep Hierarchy of Latent Variables for Generative Modeling.} In \emph{Proceedings of NeurIPS}, 2019. [0/1]
	\item Zeyuan Allen-Zhu, Yuanzhi Li, Yingyu Liang. \href{https://proceedings.neurips.cc/paper/2019/hash/62dad6e273d32235ae02b7d321578ee8-Abstract.html}{Learning and Generalization in Overparameterized Neural Networks, Going Beyond Two Layers.} In \emph{Proceedings of NeurIPS}, 2019. [0/1]
	\item Durk P. Kingma, Prafulla Dhariwal. \href{https://proceedings.neurips.cc/paper/2018/hash/d139db6a236200b21cc7f752979132d0-Abstract.html}{Glow: Generative Flow with Invertible 1x1 Convolutions.} In \emph{Proceedings of NeurIPS}, 2018. [0/2]
	\item Ricky T. Q. Chen, Yulia Rubanova, Jesse Bettencourt, David K. Duvenaud. \href{https://proceedings.neurips.cc/paper/2018/hash/69386f6bb1dfed68692a24c8686939b9-Abstract.html}{Neural Ordinary Differential Equations.} In \emph{Proceedings of NeurIPS}, 2018. [0/1]
	\item Zhitao Ying, Jiaxuan You, Christopher Morris, Xiang Ren, Will Hamilton, Jure Leskovec. \href{https://proceedings.neurips.cc/paper/2018/hash/e77dbaf6759253c7c6d0efc5690369c7-Abstract.html}{Hierarchical Graph Representation Learning with Differentiable Pooling.} In \emph{Proceedings of NeurIPS}, 2018. [0/1]
	\item Ricky T. Q. Chen, Xuechen Li, Roger B. Grosse, David K. Duvenaud. \href{https://proceedings.neurips.cc/paper/2018/hash/1ee3dfcd8a0645a25a35977997223d22-Abstract.html}{Isolating Sources of Disentanglement in Variational Autoencoders.} In \emph{Proceedings of NeurIPS}, 2018. [0/1]
	\item Yangyan Li, Rui Bu, Mingchao Sun, Wei Wu, Xinhan Di, Baoquan Chen. \href{https://proceedings.neurips.cc/paper/2018/hash/f5f8590cd58a54e94377e6ae2eded4d9-Abstract.html}{PointCNN: Convolution On X-Transformed Points.} In \emph{Proceedings of NeurIPS}, 2018. [0/1]
	\item Arthur Jacot, Franck Gabriel, Clement Hongler. \href{https://proceedings.neurips.cc/paper/2018/hash/5a4be1fa34e62bb8a6ec6b91d2462f5a-Abstract.html}{Neural Tangent Kernel: Convergence and Generalization in Neural Networks.} In \emph{Proceedings of NeurIPS}, 2018. [0/1]
	\item Ting-Chun Wang, Ming-Yu Liu, Jun-Yan Zhu, Guilin Liu, Andrew Tao, Jan Kautz, Bryan Catanzaro. \href{https://proceedings.neurips.cc/paper/2018/hash/d86ea612dec96096c5e0fcc8dd42ab6d-Abstract.html}{Video-to-Video Synthesis.} In \emph{Proceedings of NeurIPS}, 2018. [0/1]
	\item Yuanzhi Li, Yingyu Liang. \href{https://proceedings.neurips.cc/paper/2018/hash/54fe976ba170c19ebae453679b362263-Abstract.html}{Learning Overparameterized Neural Networks via Stochastic Gradient Descent on Structured Data.} In \emph{Proceedings of NeurIPS}, 2018. [0/1]
	\item Ludwig Schmidt, Shibani Santurkar, Dimitris Tsipras, Kunal Talwar, Aleksander Madry. \href{https://proceedings.neurips.cc/paper/2018/hash/f708f064faaf32a43e4d3c784e6af9ea-Abstract.html}{Adversarially Robust Generalization Requires More Data.} In \emph{Proceedings of NeurIPS}, 2018. [0/2]
	\item Shibani Santurkar, Dimitris Tsipras, Andrew Ilyas, Aleksander Madry. \href{https://proceedings.neurips.cc/paper/2018/hash/905056c1ac1dad141560467e0a99e1cf-Abstract.html}{How Does Batch Normalization Help Optimization?} In \emph{Proceedings of NeurIPS}, 2018. [0/1]
	\item Harini Kannan, Alexey Kurakin, Ian Goodfellow. \href{https://arxiv.org/abs/1803.06373}{Adversarial Logit Pairing.} In \emph{Proceedings of NeurIPS$^*$}, 2018. [0/2]
	\item Ofir Nachum, Shixiang (Shane) Gu, Honglak Lee, Sergey Levine. \href{https://proceedings.neurips.cc/paper/2018/hash/e6384711491713d29bc63fc5eeb5ba4f-Abstract.html}{Data-Efficient Hierarchical Reinforcement Learning.} In \emph{Proceedings of NeurIPS}, 2018. [0/3]
	\item Prateek Jain, Raghu Meka, Inderjit Dhillon. \href{https://papers.nips.cc/paper/2010/hash/08d98638c6fcd194a4b1e6992063e944-Abstract.html}{Guaranteed Rank Minimization via Singular Value Projection.} In \emph{Proceedings of NeurIPS}, 2010. [0/1]
	\item Hanna Wallach, David Mimno, Andrew McCallum. \href{https://papers.nips.cc/paper/2009/hash/0d0871f0806eae32d30983b62252da50-Abstract.html}{Rethinking LDA: Why Priors Matter.} In \emph{Proceedings of NeurIPS}, 2009. [0/4]
	\item Geoffrey E. Hinton, Russ R. Salakhutdinov. \href{https://papers.nips.cc/paper/2009/hash/31839b036f63806cba3f47b93af8ccb5-Abstract.html}{Replicated Softmax: an Undirected Topic Model.} In \emph{Proceedings of NeurIPS}, 2009. [0/1]
	\item Daniel J. Hsu, Sham M. Kakade, John Langford, Tong Zhang. \href{https://papers.nips.cc/paper/2009/hash/67974233917cea0e42a49a2fb7eb4cf4-Abstract.html}{Multi-Label Prediction via Compressed Sensing.} In \emph{Proceedings of NeurIPS}, 2009. [0/1]
	\item Youngmin Cho, Lawrence Saul. \href{https://papers.nips.cc/paper/2009/hash/5751ec3e9a4feab575962e78e006250d-Abstract.html}{Kernel Methods for Deep Learning.} In \emph{Proceedings of NeurIPS}, 2009. [0/1]
	\item Kurt Miller, Michael Jordan, Thomas Griffiths. \href{https://papers.nips.cc/paper/2009/hash/437d7d1d97917cd627a34a6a0fb41136-Abstract.html}{Nonparametric Latent Feature Models for Link Prediction.} In \emph{Proceedings of NeurIPS}, 2009. [0/3]
	\item Ian Goodfellow, Honglak Lee, Quoc Le, Andrew Saxe, Andrew Ng. \href{https://papers.nips.cc/paper/2009/hash/428fca9bc1921c25c5121f9da7815cde-Abstract.html}{Measuring Invariances in Deep Networks.} In \emph{Proceedings of NeurIPS}, 2009. [0/1]
	\item Vinod Nair, Geoffrey E. Hinton. \href{https://papers.nips.cc/paper/2009/hash/6e7b33fdea3adc80ebd648fffb665bb8-Abstract.html}{3D Object Recognition with Deep Belief Nets.} In \emph{Proceedings of NeurIPS}, 2009. [0/1]
	\item Martin Zinkevich, John Langford, Alex Smola. \href{https://papers.nips.cc/paper/2009/hash/b55ec28c52d5f6205684a473a2193564-Abstract.html}{Slow Learners are Fast.} In \emph{Proceedings of NeurIPS}, 2009. [0/1]
	\item Ryan Mcdonald, Mehryar Mohri, Nathan Silberman, Dan Walker, Gideon Mann. \href{https://papers.nips.cc/paper/2009/hash/d81f9c1be2e08964bf9f24b15f0e4900-Abstract.html}{Efficient Large-Scale Distributed Training of Conditional Maximum Entropy Models.} In \emph{Proceedings of NeurIPS}, 2009. [0/1]
	\item Corinna Cortes, Mehryar Mohri, Afshin Rostamizadeh. \href{https://papers.nips.cc/paper/2009/hash/e7f8a7fb0b77bcb3b283af5be021448f-Abstract.html}{Learning Non-Linear Combinations of Kernels.} In \emph{Proceedings of NeurIPS}, 2009. [0/1]
	\item Laurent Jacob, Jean-philippe Vert, Francis Bach. \href{https://papers.nips.cc/paper/2008/hash/fccb3cdc9acc14a6e70a12f74560c026-Abstract.html}{Clustered Multi-Task Learning: A Convex Formulation.} In \emph{Proceedings of NeurIPS}, 2008. [0/1]
	\item Kamalika Chaudhuri, Claire Monteleoni. \href{https://papers.nips.cc/paper/2008/hash/8065d07da4a77621450aa84fee5656d9-Abstract.html}{Privacy-preserving logistic regression.} In \emph{Proceedings of NeurIPS}, 2008. [0/3]
	\item Byron M. Yu, John P. Cunningham, Gopal Santhanam, Stephen Ryu, Krishna V. Shenoy, Maneesh Sahani. \href{https://papers.nips.cc/paper/2008/hash/ad972f10e0800b49d76fed33a21f6698-Abstract.html}{Gaussian-process factor analysis for low-dimensional single-trial analysis of neural population activity.} In \emph{Proceedings of NeurIPS}, 2008. [0/3]
	\item Ilya Sutskever, Geoffrey E. Hinton, Graham W. Taylor. \href{https://papers.nips.cc/paper/2008/hash/9ad6aaed513b73148b7d49f70afcfb32-Abstract.html}{The Recurrent Temporal Restricted Boltzmann Machine.} In \emph{Proceedings of NeurIPS}, 2008. [0/1]
	\item Wenyuan Dai, Yuqiang Chen, Gui-rong Xue, Qiang Yang, Yong Yu. \href{https://papers.nips.cc/paper/2008/hash/0060ef47b12160b9198302ebdb144dcf-Abstract.html}{Translated Learning: Transfer Learning across Different Feature Spaces.} In \emph{Proceedings of NeurIPS}, 2008. [0/3]
	\item Yishay Mansour, Mehryar Mohri, Afshin Rostamizadeh. \href{https://papers.nips.cc/paper/2008/hash/0e65972dce68dad4d52d063967f0a705-Abstract.html}{Domain Adaptation with Multiple Sources.} In \emph{Proceedings of NeurIPS}, 2008. [0/1]
	\item Sham M. Kakade, Karthik Sridharan, Ambuj Tewari. \href{https://papers.nips.cc/paper/2008/hash/5b69b9cb83065d403869739ae7f0995e-Abstract.html}{On the Complexity of Linear Prediction: Risk Bounds, Margin Bounds, and Regularization.} In \emph{Proceedings of NeurIPS}, 2008. [0/1]
	\item Francis Bach. \href{https://papers.nips.cc/paper/2008/hash/a4a042cf4fd6bfb47701cbc8a1653ada-Abstract.html}{Exploring Large Feature Spaces with Hierarchical Multiple Kernel Learning.} In \emph{Proceedings of NeurIPS}, 2008. [0/1]
	\item Ijaz Akhter, Yaser Sheikh, Sohaib Khan, Takeo Kanade. \href{https://papers.nips.cc/paper/2008/hash/dc82d632c9fcecb0778afbc7924494a6-Abstract.html}{Nonrigid Structure from Motion in Trajectory Space.} In \emph{Proceedings of NeurIPS}, 2008. [0/1]
	\item Prateek Jain, Brian Kulis, Inderjit Dhillon, Kristen Grauman. \href{https://papers.nips.cc/paper/2008/hash/aa68c75c4a77c87f97fb686b2f068676-Abstract.html}{Online Metric Learning and Fast Similarity Search.} In \emph{Proceedings of NeurIPS}, 2008. [0/1]
	\item Duy Nguyen-tuong, Jan Peters, Matthias Seeger. \href{https://papers.nips.cc/paper/2008/hash/01161aaa0b6d1345dd8fe4e481144d84-Abstract.html}{Local Gaussian Process Regression for Real Time Online Model Learning.} In \emph{Proceedings of NeurIPS}, 2008. [0/1]
	\item Lester Mackey. \href{https://papers.nips.cc/paper/2008/hash/85d8ce590ad8981ca2c8286f79f59954-Abstract.html}{Deflation Methods for Sparse PCA.} In \emph{Proceedings of NeurIPS}, 2008. [0/1]
\end{itemize}


\end{document}